\documentclass{article}

\usepackage{times}
\usepackage{graphicx} 
\usepackage{subfigure} 

\usepackage{natbib}

\usepackage{algorithm}
\usepackage{algorithmic}

\usepackage{amsmath,amsthm,amssymb}
\usepackage{commath}
\usepackage{enumitem}

\newcommand{\xhdr}[1]{\vspace{1.7mm}\noindent{{\bf #1.}}}

\DeclareMathOperator{\pr}{Pr}
\newcommand\erm{\mathrm{e}}
\newcommand\trans{^{\mathsf{\top}}}

\usepackage{Definitions}

\usepackage[accepted] {icml2017-2}

\icmltitlerunning{Fake News Mitigation via Point Process Based Intervention}
\begin{document} 
\setlength\parindent{12pt}
\onecolumn
\icmltitle{Fake News Mitigation via Point Process Based Intervention }

\begin{icmlauthorlist}
\icmlauthor{Mehrdad Farajtabar}{cse}
\icmlauthor{Jiachen Yang}{cse}
\icmlauthor{Xiaojing Ye}{gsu}
\icmlauthor{Huan Xu}{isye}
\icmlauthor{Rakshit Trivedi}{cse}
\icmlauthor{Elias Khalil}{cse} 
\icmlauthor{Shuang Li}{isye}
\icmlauthor{Le Song}{cse}
\icmlauthor{Hongyuan Zha}{cse}
\end{icmlauthorlist}

\icmlaffiliation{cse}{School of Computational Science and Engineering, Georgia Tech.}
\icmlaffiliation{gsu}{Department of Mathematics and Statistics, Georgia State University.}
\icmlaffiliation{isye}{School of Industrial and Systems Engineering, Georgia Tech.
The paper has been accepted at International Conference in Machine Learning (ICML), 2017}

\icmlcorrespondingauthor{Mehrdad Farajtabar}{mehrdad@gatech.edu}


\vskip 0.3in



\printAffiliationsAndNotice 

\begin{abstract}
We propose the first multistage intervention framework that tackles fake news in social networks by combining reinforcement learning with a point process network activity model.
The spread of fake news and mitigation events within the network is modeled by a multivariate Hawkes process with additional exogenous control terms.
By choosing a feature representation of states, defining mitigation actions and constructing reward functions to measure
the effectiveness of mitigation activities, we map the problem of fake news mitigation into the reinforcement learning framework. We develop a policy iteration method unique to the multivariate networked point process,
with the goal of optimizing the actions for maximal total reward under budget constraints.
Our method shows promising performance in real-time intervention experiments on a Twitter network to mitigate a surrogate fake news campaign, and outperforms alternatives on synthetic datasets.
\end{abstract}


\section{Introduction}
The recent proliferation of malicious fake news in social media has been a source of widespread concern.
Given that more than $62\%$ of U.S. adults turn to social media for news, with $18\%$ doing so often, fake news can have potential real-world consequences on a large scale \cite{Pew:2016a}. 
For example, within the final three months of the 2016 U.S. presidential election, news stories that favored either of the two nominees--later proved to be fake--were shared
over 37 million times on Facebook,
and over half of those who recalled seeing fake news stories believed them \cite{Allcott:2017a}.
An analysis by Buzzfeed News shows that the top 20 false election stories from hoax websites generated nearly 1.5 million more user engagement activities on Facebook than the top 20 stories from reputable major news outlets \cite{Buzzfeed:2016a}.
Therefore, there is an urgent call to develop effective strategies to mitigate the impact of fake news.

Policies to counter fake news can be categorized by the level of manual oversight and the aggressiveness of action required. Aggressively acting on fake news has various drawbacks. 
For example, Facebook's strategy allows users to report stories as potential fake news, sends these stories to fact-checking organizations, and flags them as disputed in users' newsfeed \cite{Facebook:2016a}.
Such direct action on the offending news requires a high degree of human oversight, which can be costly and slow, and also may violate civil rights.
The report-and-flag mechanism is also open to abuse by adversaries who maliciously report real news. 
Given these disadvantages, we consider an alternative strategy: optimizing the performance of real news propagation over the network, ensuring that people who are exposed to fake news are also exposed to real news, so that they are less likely to be convinced by fake news.

We face several key modeling and computational issues.
For example, how to quantify the uncertainty of user activities and news propagation within the network? 
How to measure the effect of mitigation incentives and activities?
Is it possible to steer the spontaneous user mitigation activities 
by an intervention strategy?
To address these questions, we model the temporal randomness of fake news and mitigation events (``valid news") as multivariate point processes with self and mutual excitations, in which the control incentivizes more spontaneous mitigation events by contributing to the exogenous activity of campaigner nodes.
The influence of fake news and mitigation activities is quantified using event exposure counts (i.e. the number of times that a user is exposed to fake or real news posts from other users whom she follows).


Our key contributions are as follows. We present the first formulation of fake news mitigation as the problem of optimal point process intervention in a network. The goal is to optimize the activity policy of a set of campaigner nodes to mitigate a fake news process stemming from another set of nodes.
This framework enables one to design a variety of objectives to quantify the meaning of "mitigation", such as minimizing the number of users who see fake news but were not reached by real news.
We give the first derivation of second-order statistics of random exposure counts in the non-stationary case, 
which is essential in policy evaluation and improvement.
By defining a state space for the network, formulating actions as exogenous intensity, and defining reward functions, we map the fake news mitigation problem to an optimal policy problem in a Markov decision process (MDP), which is solved by model-based least-squares temporal difference learning (LSTD) specific to the context of point processes.
Furthermore, to the best of our knowledge, we are the first to conduct a real-time point process intervention experiment.
\begin{figure*}[t]
\centering
  \begin{tabular}{c}
  \includegraphics[width=0.88\textwidth]{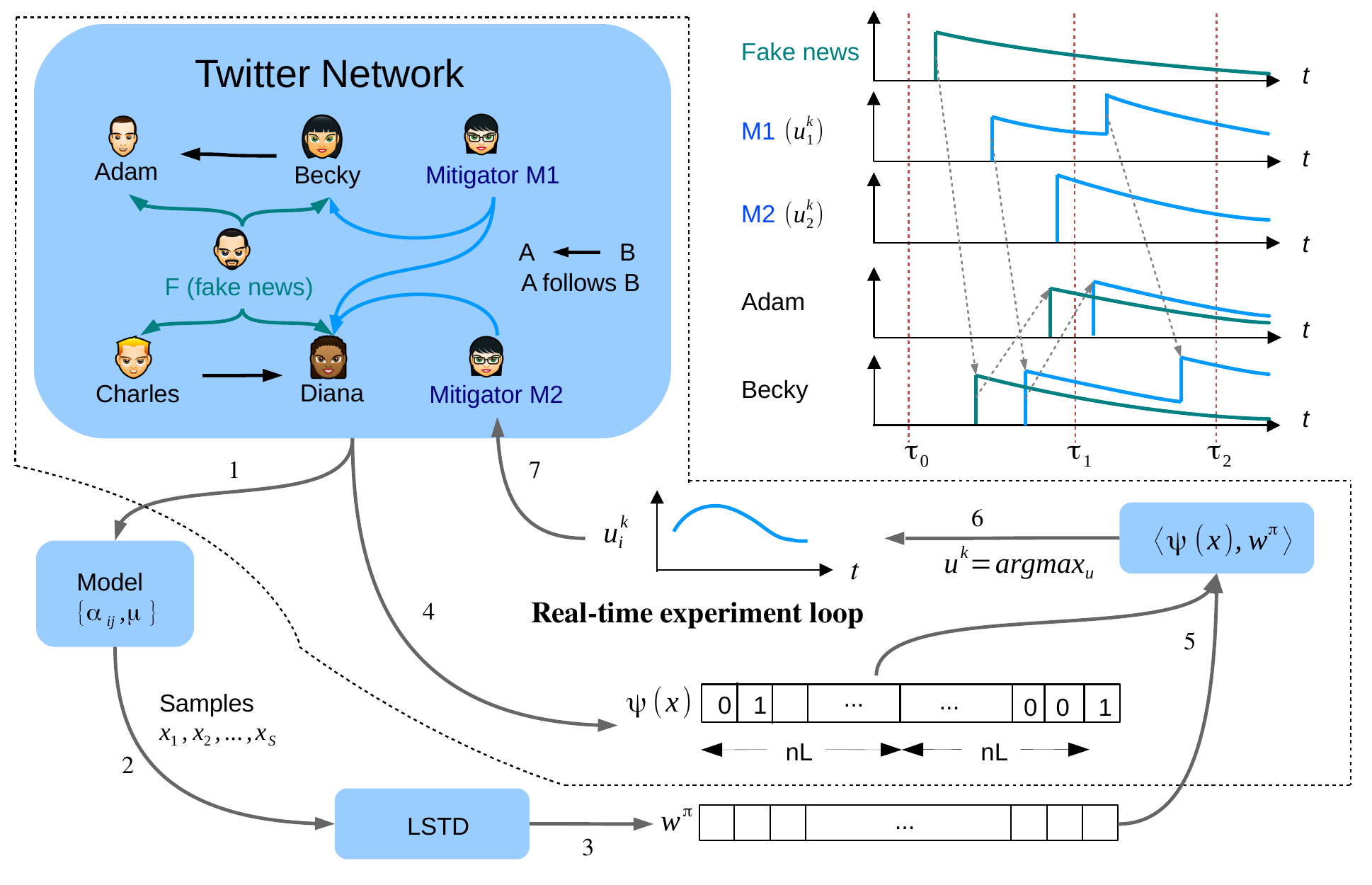}
  \end{tabular}
  \caption{The framework of point process based intervention for countering fake news. (1-3) Offline learning of value function approximation weight vector using LSTD from transition samples generated from model. (4-7) Real-time intervention loop that uses feature representation of network state to choose optimal exogenous incentive for mitigator nodes.}
  ~\label{fig:blockdiagram}
\end{figure*}

\textbf{Related work}. 
The emergence of social media as a prominent news source in the past few years raises concomitant concerns about the quality, truthfulness, and credibility of information presented \cite{Mitra:2017a}.
To reduce the amount of labor-intensive manual fact-checking, there have been research efforts devoted to building classifiers to detect factuality of information, predicting credibility level of posts, and detecting controversial information from inquiry phrases \cite{Mitra:2017a,Zeng:2016a,Zhao:2015d}.
These works mainly focused on extracting linguistic features from texts to determine the credibility of news and posts. 
Our focus in this paper, however, is to design an incentive strategy so that users can spontaneously take action to promote real news, in opposition to a real-world fake news epidemic.

Point process models have been recently used to model activities in networks~\cite{farajtabar2015coevolve,parikh2012conjoint,hosseini2017recurrent,karimi2016smart, xiao2017wasserstein}. More especially Hawkes process~\cite{Hawkes:1971a} is a class of self- and mutually exciting point processes that has been applied to variety of problems in social networks including cascade modeling~\cite{zarezade2015correlated}, reliability of crowd generated data~\cite{tabibian2016distilling}, social media popularity~\cite{rizoiuexpecting}, community detection~\cite{tran2015netcodec}, causal inference~\cite{xu2016learning}, linguistic influence~\cite{guo2015bayesian}, and change point detection in social networks~\cite{li2016detecting}.

Steering user activities by adding external incentives to the exogenous intensity of Hawkes processes was first considered in \cite{Farajtabar:2014a}. 
In \cite{Farajtabar:2016b}, a multistage campaigning method to optimally distribute incentive resources based on dynamic programming was developed. 
In these previous works, objective functions were designed using expected values of exposure counts rather than the stochastic exposure process, which may reduce the accuracy of solutions. 
Furthermore, it faced the demanding problem of computing the cost-to-go using the Hawkes model, while we address this using linear function approximation.
For stationary Hawkes processes, second order statistics was derived in \cite{Bacry:2014a,Bacry:2014b}; however, it is essential to compute both first and second order statistics for Hawkes processes in the \emph{non-stationary} stages due to time sensitivity of the fake news mitigation task, and we derive it for the first time in this paper. 
Recent work has also applied methods in stochastic differential equations to the context of point processes, to find the best intensity for information guiding \cite{WangTVS:2016} and achieving highest visibility \cite{Zarezade:2016}. While these works consider networks with only a single process, our work focuses on optimizing a mitigation process with respect to a second competing process.
Finally, although our goal is related to influence maximization problems~\cite{kempe2003maximizing,bharathi2007competitive}, our point process approach is much more general and has greater temporal resolution, as it models continuous-time recurrent activity in networks, in contrast to binary discrete-time infection states in traditional influence maximization approaches.
Moreover, our framework enables one to consider a variety of objectives (not only maximization) and incorporate budget constraints.

Reinforcement learning tackles the problem of finding good policies for actions to take in MDP where exact solutions are intractable, either due to size or lack of complete knowledge. 
Large-scale policy evaluation and iteration problems can be tackled by function approximation, which reduces the solution dimension using feature vector basis \cite{Sutton:1998:IRL:551283}.
By adding control terms to a multivariate Hawkes process model of random network activities, fake news mitigation can be formulated as a policy optimization problem in an MDP.
To address the randomness of Hawkes processes, batch reinforcement learning using samples collected from the trajectory of a fixed behavior policy can be applied \cite{Antos:2007a}. 
In particular, linear Least Squares Temporal Difference (LSTD) uses a batch of samples to learn a linear approximation of the value function under a policy with provable convergence \cite{Bradtke1996}. 
This \emph{policy evaluation} step alternates with a model-based \emph{policy improvement} step in a policy iteration to arrive at successively improved policies.  

\section{Preliminaries and Problem Statement}
\label{sec:perlim}

\textbf{Multivariate Hawkes processes}.
Hawkes process is a doubly stochastic point process with self-excitations, meaning that past events increase the chance of arrivals of new events \cite{Hawkes:1971a}, and has been extensively used to model activities in social networks~\cite{farajtabar2015coevolve,linderman2014discovering, he2015hawkestopic, rizoiuexpecting, lee2016hawkes}.
Let $t_{\ell}$ be the time of the $\ell$-th event,
then the Hawkes process can be represented
by the counting process 
\begin{align}
N(t)=\sum_{t_{\ell} \leq t} h(t-t_{\ell}), 
\end{align}
that tracks the number of events up to time $t$, where
$h(t)$ is the standard Heaviside function 
such that $h(t)=1$ if $t\geq0$ and $=0$ if $t<0$.
The conditional intensity function of a point process is defined as the probability of observing an event in an infinitesimal window given the history. For Hawkes process it is given by
\begin{align}
\lambda(t) = \mu + \sum_{t_{\ell} < t} \phi(t-t_{\ell}). 
\end{align}
Here,  
$\mu\geq0$ is the exogenous (base) intensity and $\phi(t)$ is
the Hawkes kernel that describes how fast the excitement of a past
event decays. 
In this paper, we employ the standard (stationary) exponential Hawkes kernel,
i.e., $\phi(t)=\alpha \erm^{-\omega t}h(t)$ with $\omega>\alpha>0$. 
In an $n$-dimensional multivariate Hawkes process (MHP), there are $n$
such processes $N_1(t),\dots,N_n(t)$ that can also
mutually excite one another, and the conditional intensity 
$\lambda(t):=\del{\lambda_1(t),\dots,\lambda_n(t)}\trans \in\mathbb{R}_+^n$ is given by
\begin{align}
\lambda(t) = \mu + \int_0^t \Phi(t-s) \dif N(s).
\end{align}
Here, $N(t):=\del{ N_1(t),\dots,N_n(t)}\trans \in \mathbb{N}_0^n$,
$\mu:=(\mu_1,\dots,\mu_n)\trans \in\mathbb{R}_+^n$, 
and $[\Phi(t)]_{ij}=\phi_{ij}(t):=\alpha_{ij} \erm^{-\omega t}h(t)$.
We let $\Hcal(t)$ denote the filtration of $N(t)$, generated by
the $\sigma$-algebra of history $\cbr{(t_{\ell},i_{\ell}) | t_{\ell}\leq t}$ 
of this point process, where $i_{\ell}\in\cbr{1,\dots,n}$ is the
identity (node) of the $\ell$-th event. 

\textbf{Network activities}.
We model both fake news and mitigation processes as MHP over the network. 
Conceptually, MHP is a networked point process model with 
dependent dimensions (nodes), and can capture the underlying network structure and node interactions~\cite{blundell2012modelling,Xu:2016a, guo2015bayesian}.
For example, an event by one user (a node) can trigger more events at other connected users.
Define $F(t)=\del{F_1(t),\dots,F_n(t)}\trans \in \mathbb{N}_0^n$, where $F_i(t)$ counts the number of times user $i$ shares a piece of news from the fake campaign up to time $t$. 
Similarly, define $M(t)=\del{M_1(t),\dots,M_n(t)}\trans \in \mathbb{N}_0^n$ for the mitigation process.
Correspondingly, we have 2 intensity functions: $\lambda^M(t) = (\lambda^M_1(t) , \ldots, \lambda^M_n(t) )^{\top}$ and
$\lambda^F(t) = (\lambda^F_1(t) , \ldots, \lambda^F_n(t) )^{\top}$ and two sets of exogenous intensities $\mu^M$ and $\mu^F$.  

\textbf{Goal}.
Given that both $F(t)$ and $M(t)$ are modeled by the Hawkes processes, 
our goal is to find the optimal mitigation strategy 
that specifies how to adjust the exogenous intensity of a few mitigator nodes, such that an objective function (rigorously defined in sec.~\ref{sec:fakenewsmitigation}) can be maximized under budget constraints. 
To this end, we measure the influence of fake news and mitigation activities using event exposures, describe the mechanism of mitigation interventions, and quantify the effect of interventions mathematically.

\textbf{Event exposure}. 
Event exposure is a quantitative measure of campaign influence, and is represented as a counting process, 
$\Ecal(t) = \del{\Ecal_1(t),\dots,\Ecal_n(t)}\trans $. 
Here, $\Ecal_i(t)$ records the number of times user $i$ is exposed to a campaign $N(t)$ by time $t$, where the exposure count increases whenever the user or a neighbor performs an activity. 
Let $B$ be the adjacency matrix of the user network, i.e., $b_{ij} = 1$ if user $i$ follows user $j$, and assume $b_{ii} =1$ for all $i$. 
Then the exposure process is given by $\Ecal(t)=BN(t)$.
We define $\Fcal(t)=BF(t)$ and $\Mcal(t)=BM(t)$ as the fake news and mitigation exposure processes, respectively.
Note that the MHP allows cascades of mutual excitations to occur among many nodes, so that non-adjacent users can also contribute to one another's exposure counts, if there is a directed path between them.

\textbf{Intervention}. 
To maximize objectives defined for fake news mitigation in section~\ref{sec:fakenewsmitigation}, suppose we can perform intervention by incentivizing a subset of users in the $k$-th stage during time $[\tau_k,\tau_{k+1})$ to trigger real news events.
For simplicity, we consider uniform time duration 
$\tau_{k+1}-\tau_{k}=\Delta_T$ for $k=0,1,\dots$, since
generalization to nonuniform time durations is trivial.
We model the incentive by a constant intervention $u_i^k\geq0$ added to the exogenous intensity $\mu_i$ during time $[\tau_{k},\tau_{k+1})$ for each stage $k=0,1,\dots$.
The mitigation activity intensity at the $k$-th stage is
\begin{align}
\lambda^M(t) = \mu + u^k +  \int_{0}^t \Phi(t-s) \dif M(s), 
\end{align} 
for $t \in [\tau_k,\tau_{k+1})$.
Note that the intervention itself exhibits a stochastic nature: adding $u_i^k$ to $\mu_i$ is equivalent to incentivizing user $i$ to increase her activity rate, but it is still uncertain when she will perform an activity, which appropriately mimics the randomness of the real world.

\textbf{Reward function}.
For each stage $k$, let $x^k$ (defined in section~\ref{sec:state_representation}) be the state of the whole MDP that encodes all the information from previous stages, and let $u^k$ be the current control imposed at this stage.
Let $\Mcal_i^k(t;x^k,u^k):=\sum_j b_{ij}\int_{\tau_k}^t \dif M_j(s)$ be the number of times user $i$ is exposed to the mitigation campaign by time $t\in[\tau_k,\tau_{k+1})$ within stage $k$, and define $\Fcal_i^k(t;x^k, u^k)$ similarly for the fake news exposure process.
The reward function $R(x^k, u^k)$ can then be designed as a composite function of $\Mcal$ and $\Fcal$ (section~\ref{sec:fakenewsmitigation}).

\textbf{Problem statement}. 
By observing the counting process in previous stages (summarized in
a sequence of $x^k$) and
taking the future uncertainty into account, the control problem
is to design a policy  $\pi$ 
such that the controls $u^k=\pi(x^k)$ can maximize the total discounted objective
\begin{align}
\EE [ \sum_{k=0}^{\infty}  \gamma^k R^k], \end{align} 
where $\gamma\in(0,1]$ is the discount rate and $R^k$ is the observed reward
at stage $k$.
In addition, we may have constraints on the amount of control, such as a budget constraint on the sum of all interventions to 
users at each stage, or a cap over the amount of intensity a user 
can handle.  
A feasible set or an action space over which we find the best intervention 
is represented as
\begin{align}
U_k := \cbr{u \in \mathbb{R}^n | u\trans c^k \leq C_k, 0 \leq u \leq \alpha^k}.
\end{align}
Here,  $c_i^k$ 
is the price per unit increase of exogenous intensity of user $i$ 
and $C_k \in \mathbb{R}_+ $ is the total budget at stage $k$. 
Also, $\alpha^k_i$ is the cap on the amount of activities of the user $i$. 



\section{Proposed Method}
In this section, we present the formulation of reward functions in terms of event exposures of fake news and mitigation activities. 
Then we derive the key statistics of the MHP required for reward function evaluation, followed by the policy iteration scheme to find the optimal intervention.  

\subsection{Fake news mitigation}
\label{sec:fakenewsmitigation}
As we discussed above, the total reward of policy $\pi$ is defined by
the value function
\begin{align}\label{eq:valfn}
V^{\pi}(x^0)= \EE \sbr[3]{\sum_{k=0}^{\infty} \gamma^k R^k \sVert[3] x^0}
\end{align}
for the initial state $x^0$ of fake and mitigation processes,
where the observed reward $R$ quantifies the effect of mitigation activities $M(t)$
in each stage and $\gamma\in(0,1]$ is the discount rate.
We consider two types of reward functions $R(x,u)$: 


\noindent
1) \emph{Correlation Maximization}:
One possible way is to require correlation between mitigation exposures and fake news exposures: people exposed more to fake news should also be exposed more to the true news, so that they are less likely to believe completely in fake news. 
Therefore, we can design the reward function $R$ in stage $k$ to be:
\begin{equation*} 
R(x^k,u^k)= \frac{1}{n} \Mcal^k(\tau_{k+1};x^k,u^k)\trans \Fcal^k(\tau_{k+1};x^k,u^k).
\end{equation*}
2) \emph{Difference Minimization}:
Suppose the goal is to minimize the number of unmitigated fake news events, then we can form a reward function $R$ in stage $k$ as the least squares of unmitigated numbers:
\begin{equation*}\label{eq:AUM}
R(x^k,u^k) = \frac{-1}{n} \enVert{\Mcal^k(\tau_{k+1};x^k,u^k)-\Fcal^k(\tau_{k+1};x^k,u^k)}^2
\end{equation*}
These are two candidate reward functions in the MHP-MDP context, among others.
To solve the policy optimization problem $\argmax_{\pi}V^{\pi}(x^0)$ for $V^{\pi}$ defined
in \eqref{eq:valfn}, we need to evaluate the value function $V^{\pi}$ for any given policy $\pi$,
which requires the first and second order statistics (moments) of any multivariate Hawkes processes $N(t)$,
as we derive next.

\subsection{Second order statistics of non-stationary MHP}
\label{sec:2nd-order}
For an $n$-dim MHP $N(t)$ with 
standard exponential kernel $\Phi(t)$, the following proposition
provides closed-form solution of the mean intensity
$\eta(t):=\EE[\lambda(t)]$ for both constant and time-varying exogenous intensity $\mu(t)$:
\begin{proposition}[Theorem 3 \cite{Farajtabar:2014a,Farajtabar:2016b}]
\label{prop:eta}
Let $N(t)$ be an $n$-dimensional MHP defined in sec.~\ref{sec:perlim} 
with exogenous intensity $\mu(t)$ and
Hawkes kernel $\Phi(t)=A \erm^{-\omega t}h(t)$,
then the mean intensity $\eta(t)$ is given by 
\begin{equation}\label{eq:eta}
\eta(t)=\sbr[1]{\erm^{(A-\omega I)t}+\omega (A-\omega I)^{-1}\del[1]{\erm^{(A-\omega I)t} - I} }\mu(t).
\end{equation}
\end{proposition}
Let $\Lambda(t)=\int_{0}^t\lambda(s)\dif s$
be the compensator of $N(t)$, 
then by the Doob-Meyer decomposition theorem, $N(t)-\Lambda(t)$ is a zero mean
martingale. This implies that the first order statistics $\EE[N(t)]$
can be obtained by $\EE[N(t)]=\EE[\Lambda(t)]=\EE[\int_0^t\lambda(s)\dif s]
= \int_0^t \EE[\lambda(s)]\dif s=\int_0^t\eta(s)\dif s$
using eq.~\eqref{eq:eta}.

To evaluate the reward function $R$ defined previously, we need to derive
second order statistics of multivariate Hawkes process $N(t)$ in its non-stationary
stage. The following theorem states the key ingredients for
the second order statistics. The proof is provided in the appendix.
\begin{theorem}\label{thm:secondorder}
Let $N(t)$ be an $n$-dim MHP with exogenous intensity $\mu$ and
Hawkes kernel $\Phi$ defined in sec.~\ref{sec:perlim}, 
then the second order statistics of $N(t)$ for $t,t'\geq0$ is given by
\begin{equation}
\begin{split}
\EE\sbr{\dif N(t)\dif N(t')\trans  } &  = G(t',t)\trans \Sigma(t')\dif t\dif t' + \\ 
\delta(t-t')&\Sigma(t')\dif t\dif t'+\eta(t)\eta(t')\trans \dif t\dif t'
\end{split}
\end{equation}
where $\eta(t)=\EE[\lambda(t)]$ is given in \eqref{eq:eta},
$\Sigma(t)=\diag([\eta_i(t)])$ is diagonal, and $G$ is the unique solution of
\begin{equation}
G(t',t)=G(t',t) \ast \Phi(t) + \Phi(t-t') - \delta(t-t')I.
\end{equation}
Moreover $G(t',t)\trans \Sigma(t')=\Sigma(t)G(t,t')$ for all $t,t'\geq0$.
\end{theorem}
Based on Theorem \ref{thm:secondorder}, we
can compute second order statistics such as $\EE[N_i(t)N_j(t')]$ for all
$i,j$ and $t,t'\geq0$.

\subsection{State Representation}
\label{sec:state_representation}
Hawkes process is non-Markovian and one needs complete knowledge of the history to characterize the entire process. 
However, when the standard exponential kernel $\Phi(t,s)=Ae^{-\omega (t-s)}h(t-s)$ is employed, the effect of history up to time $\tau_k$ 
on the future $t>\tau_k$ can be cleverly summarized by one scalar per dimension~\cite{simma2012modeling,Farajtabar:2016b}. 
For $1 \le i \le n$, define 
$
y^k_i := \lambda^{k-1}_i(\tau_{k}) - u^{k-1}_i - \mu_i
$,
(and $y_0^i = 0$ by convention), then the intensity due to events of all previous $k$ stages can be written as
$
\int_{0}^{\tau_k} A  e^{-\omega(t-s)} \, \dif N(s) = y^k e^{-\omega(t-\tau_k)}
$.
In other words, $y^k$ is sufficient to encode the information of activities 
in the past $k$ stages that are relevant to future. 
Note that we have two separate $y^k_M$ and $y^k_F$ to track the dynamics of both mitigation and fake processes.

In order to tackle objectives over multiple stages, we add aggregated number of events at $L$ previous $\Delta_f$-time intervals 
over all dimensions. 
Define a vector 
$z^k \in \RR^{n L}$ where
$
z^k_{(l-1)n+i} = \int_{\tau_k-l\Delta_f}^{\tau_k- (l-1)\Delta_f}
\, \dif N_i(s)
$ 
for $1 \le i \le n$ and $1 \le l \le L$. 
In other words, $z^k_{(l-1)n+i}$ records the number of events of $i$-th dimension in the $l$-th interval of length $\Delta_f$ 
prior to time $\tau_k$.
For example, choosing $\Delta_f = \Delta_T$ and setting $L=2$ means that events from the two most recent stages are counted.
Similarly, we have two separate $z^k_M$ and $z^k_F$ corresponding to the two processes.
Now, the state vector $x^k \in \RR^{2nL + 2n}$ is the concatenation of the above four vectors $x^k = [y^k_M ;  y^k_F ; z^k_M ; z^k_F]$.

\subsection{Least Squares Temporal Difference}
\label{sec:policyiteration}
The optimal value function satisfies the Bellman equation: 
\begin{align}
V^{\pi} (x) =\EE[ R (x,\pi(x))] + \gamma \EE [ V^{\pi} (x')],
\end{align}
where $x'$ is the next state after taking action based on policy $\pi$ at state $x$.
Least squares temporal difference learning (LSTD) is a sample-efficient procedure for policy evaluation, which subsequently facilitates policy improvement. The value function is approximated by 
$\hat{V}^{\pi}(x) = \sum_{d=1}^D w^{\pi}_d \psi_d(x)$, 
where $\psi_d $ is the $d$-th feature of state $x$ and $w^{\pi}_d $ is its coefficient for policy $\pi$.
This can be compactly represented as $\hat{V}^{\pi}(x) = \psi(x)^{\top} w^{\pi}$,
where $\psi(x) = (\psi_1(x) , \ldots, \psi_D(x) )^{\top}$.
The following presents our choice of features and the policy evaluation and improvement steps of LSTD(0)~\cite{Sutton:1998:IRL:551283}.

\textbf{Features}.
The number of events in a few recent consecutive intervals of point processes have been used as a reliable feature to parameterize point processes~\citep{parikh2012conjoint,qin2015auxiliary,lian2015multitask}. Following their work we take $L$ prior intervals of length $\Delta_f$ for each dimension of the fake news process and record the number of events in that period as one feature. 
 $\psi^k_{(l-1)n+i} = z^k_{(l-1)n+i}$ for $1 \le i \le n$ and $1 \le l \le L$. 
This will count for $nL$ features. Similarly we take $nL$ features from the mitigation process. Finally,  we add a last feature $\psi^k_{2nL+1} = 1$ as the bias term.  Therefore, $\psi^k = [z^k_M ; z^k_F; 1]$ and the feature space has dimension $D = 2nL + 1$.

\begin{algorithm}[tb]
\caption{LSTD policy iteration in point processes}
\label{alg:lspi}
\begin{algorithmic}
   	\STATE {\bfseries Input:} set of samples $\Scal$, feature $\psi(\cdot)$, discount $\gamma$
   	\REPEAT
   		\STATE Initialize $A^{\pi} = 0$ and $b^{\pi} = 0$.
   		\FOR{{\bfseries each} state $x \in \Scal$}
   			\STATE $A^{\pi} \leftarrow A^{\pi} + \psi(x)(\psi(x) - \gamma \psi(x'))\trans$
            \STATE $b^{\pi} \leftarrow b^{\pi} + \psi(x) r^{\pi}$
   		\ENDFOR
        \STATE $w^{\pi} \leftarrow (A^{\pi})^{-1} b^{\pi}$
        \FOR{{\bfseries each} state $x \in \Scal$}
        	\STATE $\pi(x) \leftarrow \underset{u}{\argmax} \lbrace \EE[ R(x,u) ]  + \gamma \EE[ V^{\pi}(x') | u, w^{\pi} ] \rbrace $
        \ENDFOR
   	\UNTIL{$\lVert \Delta w^{\pi} \rVert < 0.1$} 
    \STATE {\bfseries return} $w^{\pi}$
\end{algorithmic}
\end{algorithm}

{\bf Policy Evaluation.}
Substituting the approximation into the Bellman equation, we have:
\begin{align}
\label{eq:bellman-with-feature}
\psi(x)^{\top} w^{\pi} = \EE[ R (x,\pi(x))]  + \gamma  \EE[\psi(x')^{\top}]  w^{\pi}.
\end{align}
To find the best fit of $w^{\pi}$ we have to consider all possible $x$; however, 
since the state space is infinite-dimensional, enumerating all states is impossible and we utilize a set $\Scal$ of samples $\mathcal{S} = \{ x_1, \ldots, x_S \}$. 

Let $\psi(x_s) = \psi_s \in \RR^D$,
 $\EE [ \psi(x'_s) ] = \psi'_s \in \RR^D$, and $r^{\pi}_s = \EE[ R (x_s,\pi(x_s))] \in \RR$. Then  define matrices of current features $\Psi = [\psi_1^{\top}; \ldots; \psi_S^{\top}]^{\top} \in \RR^{S \times D}$ and next features 
 $\Psi' = [ {\psi'}_1^{\top}; \ldots ; {\psi'}_S^{\top} ]^{\top} \in \RR^{S \times D}$, the rewards $r^{\pi} = [r^{\pi}_1, \ldots, r^{\pi}_S]^{\top} \in \RR^S$, and the sample value functions as
  $v^{\pi} = [V^{\pi}(x_1), \ldots, V^{\pi}(x_S)]^{\top} \in \RR^S$.
Appendix \ref{sec:app:policy-improve} presents how we leverage the first and second order statistics of Hawkes process to find $\EE[ R (x,u)]$ and $\EE[V^{\pi} (x')]$.
Given the above definition, the Bellman optimality of eq.~\eqref{eq:bellman-with-feature} can be written in matrix format:
 \begin{align}
v^{\pi} = \Psi w^{\pi} = r^{\pi}  + \gamma \Psi' w^{\pi} \triangleq T^{\pi} v^{\pi},
\end{align}
where $T^{\pi}$ is the Bellman optimality operator.
A way to find a good estimate is to force the approximate value function
to be a fixed point of the optimality equation under the Bellman operator, \emph{i.e.},
$
T^{\pi}   \hat{v}^{\pi} \approx \hat{v}^{\pi}. 
$ \cite{Lagoudakis:2003a}.
For that, the fixed point has to lie in the space of approximate value functions, spanned by the basis functions $\Psi$. $ \hat{v}^{\pi} $ lies in that space by definition, but $T^{\pi} \hat{v}^{\pi}$ may have an orthogonal component and must be projected. This is achieved by the orthogonal projection operator $(\Psi(\Psi^{\top}\Psi)^{-1}\Psi^{\top})$. Therefore the approximate value function $\hat{v}^{\pi}$ must be invariant under one application of the Bellman operator $T^{\pi} $ followed by orthogonal projection:
\begin{align}
\hat{v}^{\pi}  = \Psi(\Psi^{\top}\Psi)^{-1}\Psi^{\top} (T^{\pi}   \hat{v}^{\pi} ).  
\end{align}
By substituting the linear approximation $\Psi  w^{\pi} = v^{\pi}$ into the above equation and some manipulations, we get a $D \times D$ linear systems of equations $A^{\pi} \omega^{\pi} = b^{\pi}$, where $A^{\pi} = \Psi^{\top}  (  \Psi   - \gamma \Psi')$ and $b^{\pi} =  \Psi^{\top} r^{\pi}$, and whose solution is the fitted coefficients $w^{\pi}$. It has been shown that the estimated $w^{\pi}$ converges to the best $w^*$ as the available number of samples tends to infinity \cite{Bradtke1996}. Appendix~\ref{sec:policy-eval} presents a detailed derivation.

\begin{algorithm}[tb]
\caption{Real-time fake news mitigation}
\label{alg:mitigation}
\begin{algorithmic}
   	\STATE {\bfseries Input:} network $A$, learned $w^{\pi}$, feature $\psi(\cdot)$, discount $\gamma$
   	\REPEAT
   		\STATE Observe state $x$ of the network activities
        \STATE $u = \argmax_a \lbrace \EE[R(x,a)] + \gamma \EE[ V^{\pi}(x') | a, w^{\pi} ] \rbrace $
        \STATE Add $u$ to base exogenous intensity $\mu$ and generate mitigation event times $\lbrace t_i \rbrace$ using point process model
        \STATE Create posts at times $\lbrace t_i \rbrace$ using campaigner accounts
   	\UNTIL{end of campaign}
\end{algorithmic}
\end{algorithm}

{\bf Policy Improvement.}
The second part of the algorithm implements policy improvement, i.e.,  getting an improved policy $\pi'$ via one-step look-ahead as follows:
\begin{align}
\label{eq:argmax}
\pi'(x) = \argmax_u \EE[ R (x,u)  + \gamma V^{\pi} (x') ].
\end{align}


LSTD(0) alternates between the policy improvement and policy evaluation iteratively until $w^{\pi}$ converges \cite{Bradtke1996}.
Alg.~\ref{alg:lspi} summarizes this procedure.

\textbf{LSTD in Hawkes context}.
LSTD is particularly suitable to the problem we are interested in. It learns the value function $V^\pi(x)$, and as such, policy improvement can be challenging without knowing the model. Because of this, methods that aim to learn the Q-function $Q^{\pi}(x,u)$, such as LSPI \cite{Lagoudakis:2003a}, are widely applied. The downside of Q-function based methods is that they typically require more samples than value-function based methods, because they require a large collection of state-action pairs $\lbrace (s,a) \rbrace$ for sufficient exploration of both state and action spaces. 
Yet, in our setup, learning the value function is sufficient, by writing the action-value function as
$Q^{\pi}(x,u) = \EE [ R(x,u) + V^{\pi}(x') ]$, and observing that the learned model of the multivariate Hawkes process enables analytical computation of the expectation (see Appendix \ref{sec:app:policy-improve} for details): 
\begin{equation*} 
\begin{split}
 \EE  [V^{\pi}(x')] & = \sum_{i=1}^n \sum_{l=1}^{L-1} w^{\pi}_{ln+i} z^{k-1}_{M,(l-1)n+i} 
+ w^{\pi}_{nL+ln+i} z^{k-1}_{F,(l-1)n+i}
 \\
  & +  \sum_{i=1}^{n} w_{i} \EE[z^k_{M,i}]+ w_{nL+i} \EE[z^k_{F,i}] + w^{\pi}_{2nL+1},   
 \\
 \EE [ R(x,u)] & =  \frac{1}{n} \EE[ {z^k_{M}}]^{\top} \, B^{\top}  B  \, \EE[z^k_{F} ] ,
   \quad \quad \text{ \% correlation}
 \\
  \EE [ R(x,u)] & =  - \frac{1}{n} \EE[ {z^k_{M}}^{\top} \, B^{\top}  B  \, z^k_{M} ]
   - \frac{1}{n} \EE[ {z^k_{F}}^{\top} \, B^{\top}  B  \, z^k_{F} ] \\ 
  & 
  + \frac{2}{n} \EE[ {z^k_{M}}]^{\top} \, B^{\top}  B  \, \EE[z^k_{F} ].
     \quad \quad \text{ \% difference}
\end{split}
\end{equation*}

We require much fewer samples to learn $V^{\pi}(x)$ compared to learning an approximate $Q^{\pi}(x,u)$, and in particular compared to LSPI, we avoid explicitly discretizing the continuous action space from which the action $u$ is chosen.

We further remark that the policy improvement step finds the optimal action $u$ at any state $x$  by
 computing
$\argmax_u \EE [ R(x,u) + V^{\pi}(x') ]$, where the action $u$ to be optimized appears in the calculation of both the expected current reward and the expected value at the next state.
This optimization problem is convex under our choice of reward functions and the form of the Hawkes conditional intensity.

After learning the optimal policy (implicitly defined by $w^{\pi}$ of the linearly-approximated value function) we start at the real-time intervention part. Given a state observation, we find the optimal intervention intensity by solving eq.~(\ref{eq:argmax}). Alg.~\ref{alg:mitigation} summarizes the real-time mitigation procedure.



\begin{figure*}[t]
\centering
  \begin{tabular}{cccc}
  \includegraphics[width=0.24\textwidth]{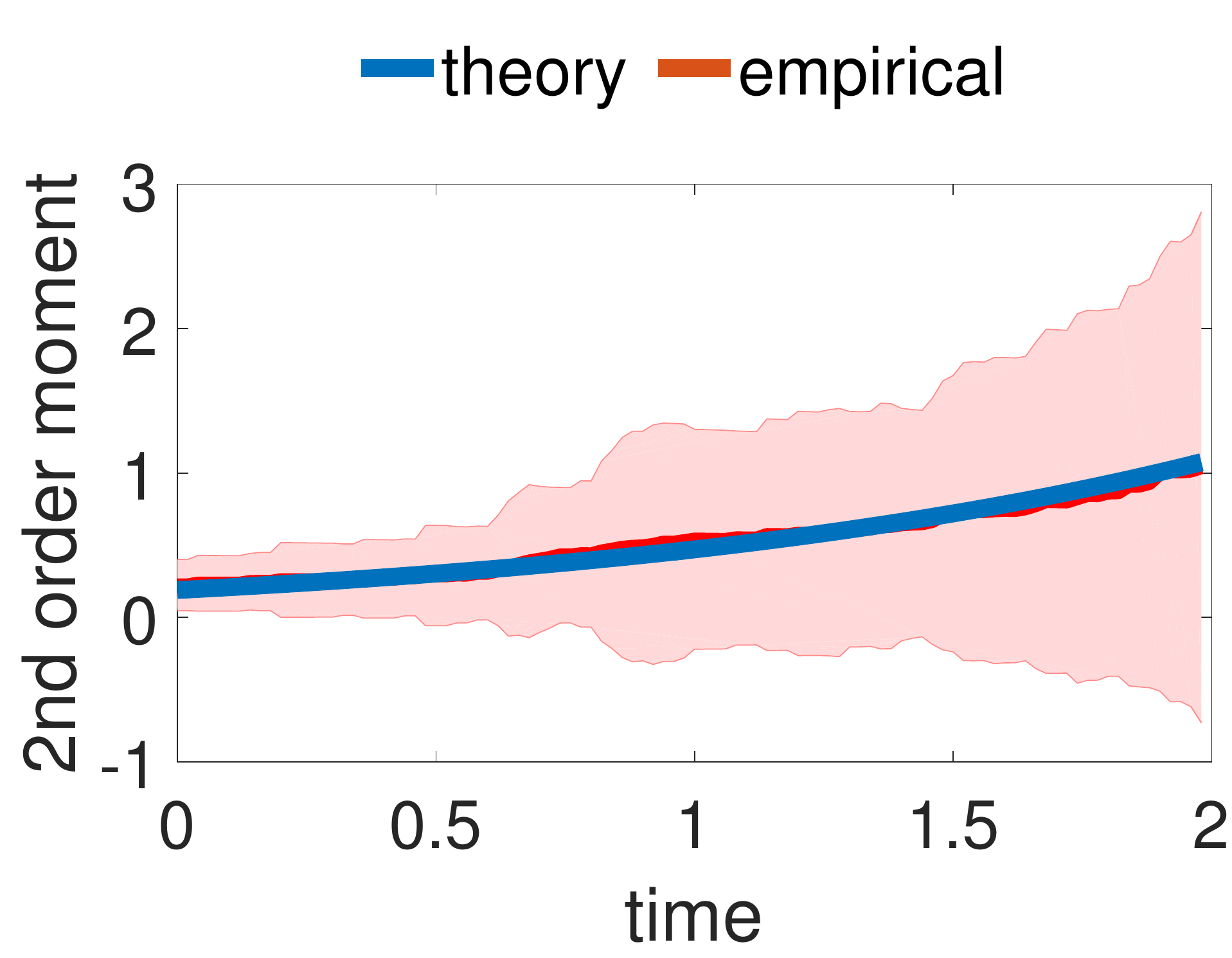}  \hspace{-3mm} &
  \includegraphics[width=0.24\textwidth]{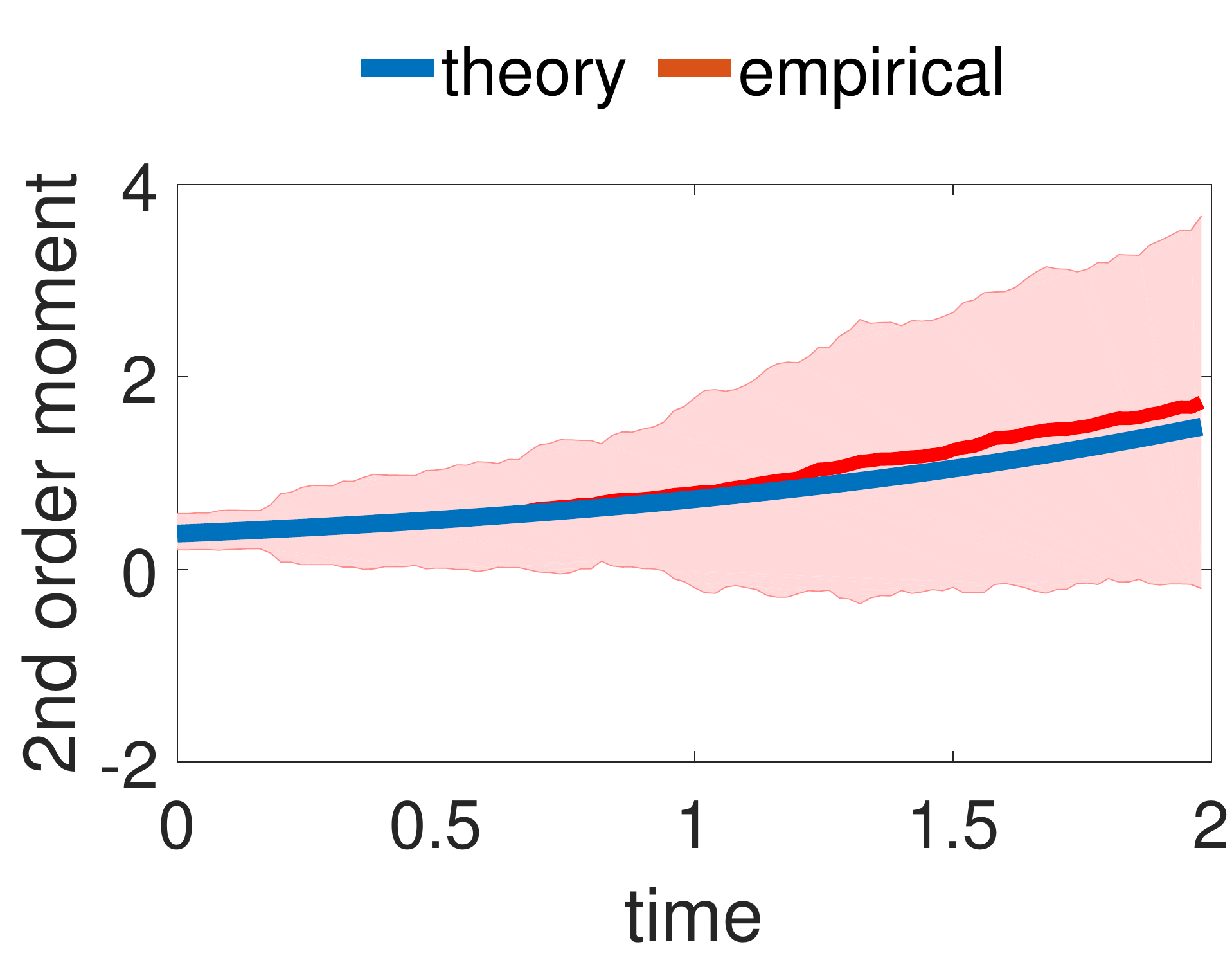} \hspace{-3mm} &
  \includegraphics[width=0.24\textwidth]{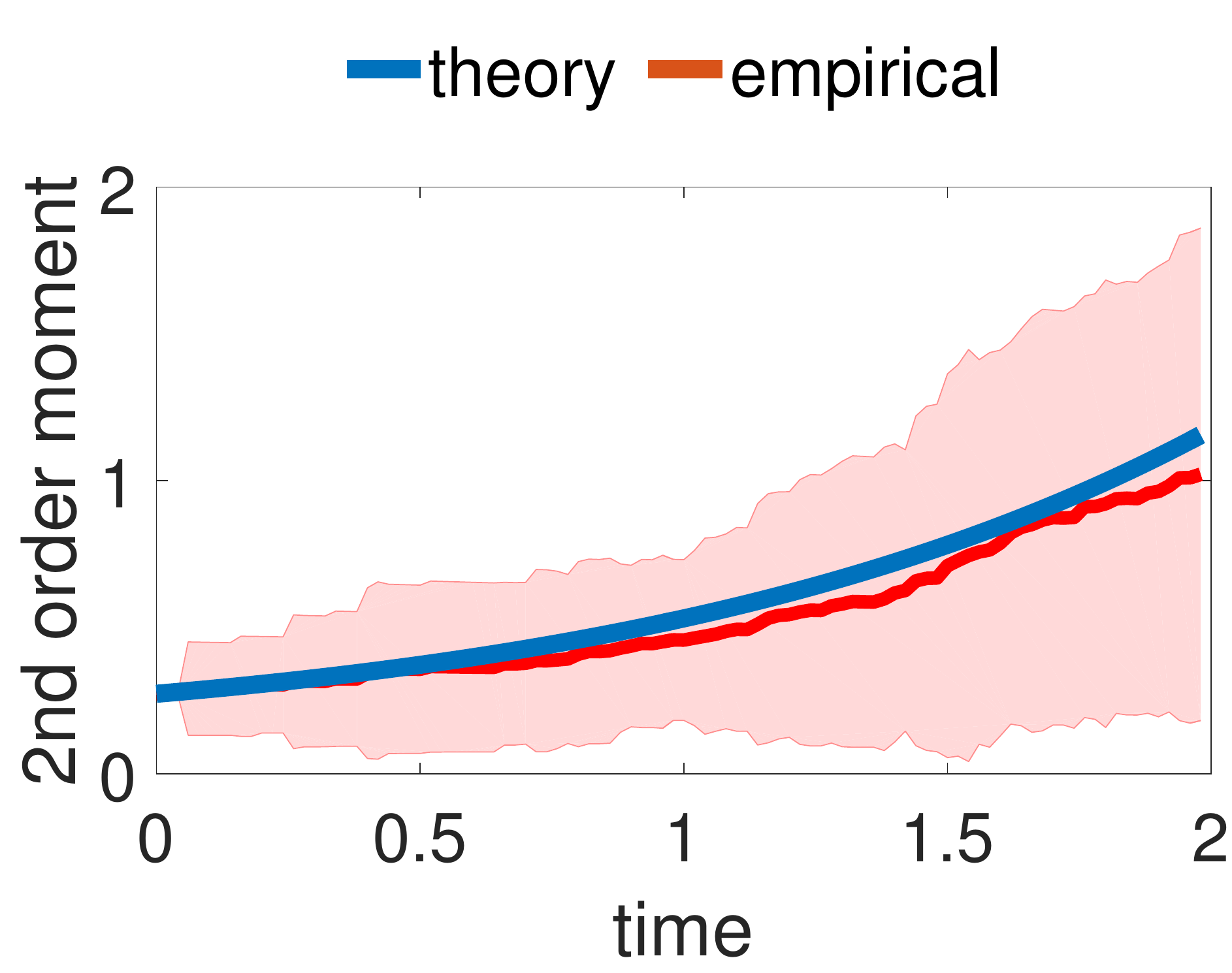} \hspace{-3mm} &
  \includegraphics[width=0.24\textwidth]{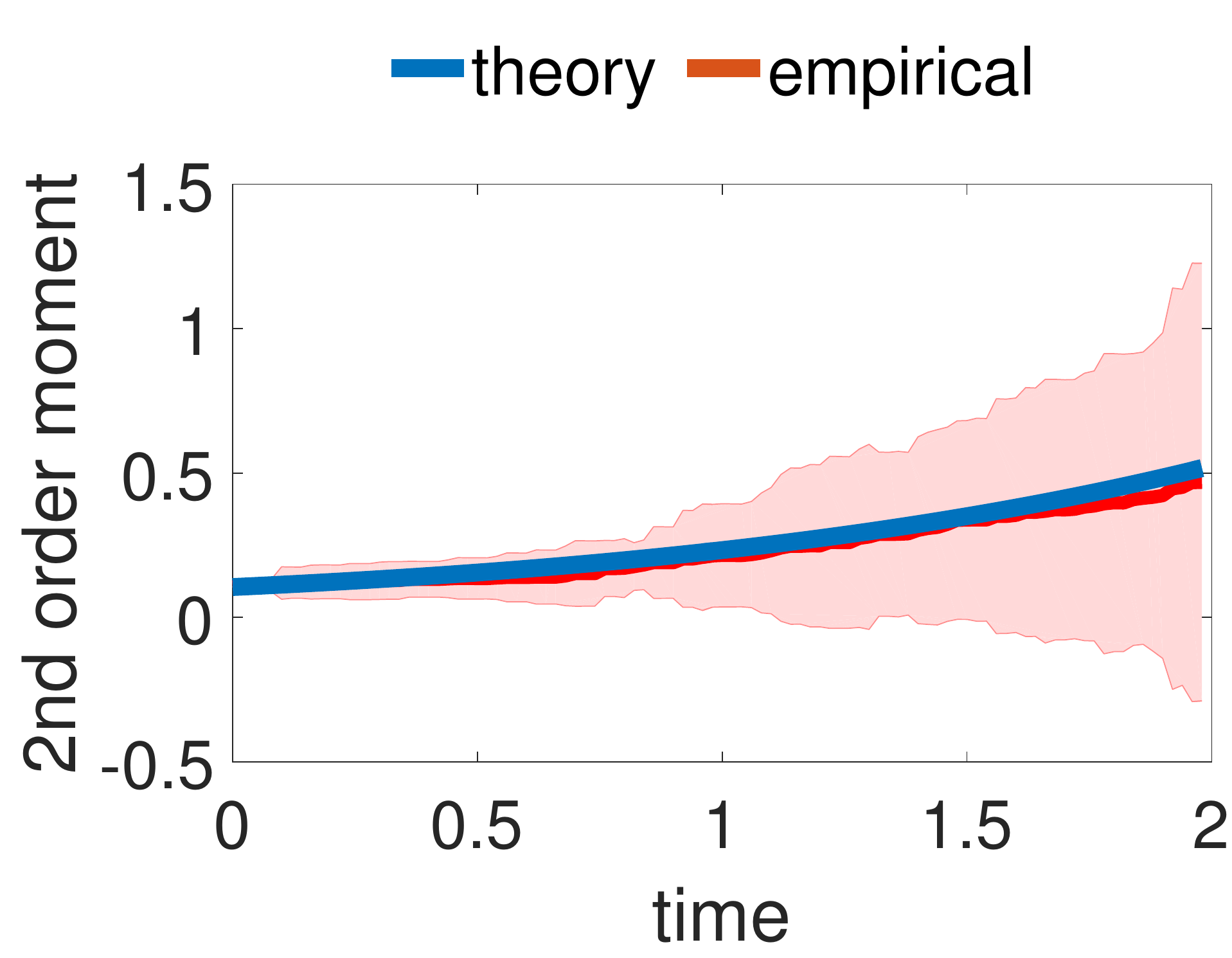}  \hspace{-3mm}
  \end{tabular}
  \caption{Empirical and theoretical second order moments of a Hawkes process, $\EE[\dif N_i(t)\dif N_j(t')]$ for 4 random pairs $(i,j)$ and $t'=0$ and varying $t$ from $0$ to $2$. }
  ~\label{fig:second-order}
\end{figure*}

\section{Experiments}
We evaluated our fake news mitigation framework by both simulated and real-time real-world experiments and show that our approach significantly outperforms several state-of-the-art methods and alternatives. 
First we verify the theoretical second order statistics in Fig.~\ref{fig:second-order}, whose proof can be found in appendix~\ref{thm:secondorder}.
Then we introduce the baseline methods against which we compare our proposed approach, and present the results of synthetic and real intervention experiments.
The measure of performance for all methods was how much total reward could be accumulated by each method, where the reward function is defined via the objective functions in section~\ref{sec:fakenewsmitigation}.
We conclude by examining convergence properties and representative power of the chosen linear features in section~\ref{sec:policyiteration}.

\subsection{Empirical validation of second order statistics}
\label{sec:second-order}
In this section, we empirically study the theoretical results of section~\ref{sec:2nd-order}. The mean and standard deviation of the empirical second order statistics averaged over 100 simulations was compared to the theoretical mean. This experiment helps to verify that it can be used in simulations to evaluate the merits of our proposed algorithm and versus the baselines. 
Fig.~\ref{fig:second-order} demonstrates the second order correlation  profile of 4 random pairs of users simulated 100 times. 
We see that the empirical average almost matches the theoretical average. 
Furthermore, it is interesting to see that the standard deviation increases with time. This is due to the aggregation of more random elements as time passes. Therefore, one should be careful with using the empirical mean without a sufficient number of random runs when the time interval is large.

\subsection{Baselines}
\label{sec:baselines}
We compared our Least-squares Temporal Difference (\textbf{LTD}) intervention procedure with the following baselines. Sections~\ref{sec:synthetic_experiments} and~\ref{sec:real_experiments} present experimental procedures and results.

1) \textbf{CEC}~\cite{Farajtabar:2016b}: This is a recent network intervention algorithm based on point processes.
It formulates a dynamic programming problem,
 \begin{align}
 V^k(x) =  \max_{u^k} \EE[ R (x^k,u^k)  + \gamma V^{\pi} (x^{k+1}) ],
 \end{align}
 and uses approximate look-ahead dynamic programming implemented via Certainly Equivalence Control (CEC)~\cite{bertsekas1995dynamic} to find the optimum intervention.

2) \textbf{OPL}~\cite{Farajtabar:2014a}: An open loop dynamic programming control based on convex optimization that finds the best intervention for all stages in one shot:
 \begin{align}
  \max_{u^1, u^2, \ldots, u^K}  \EE[ \sum_k \gamma^{k}  R (x^k,u^k)  ].
 \end{align}
Open Loop (OPL) is an important baseline, because comparing it against closed-loop strategies like CEC and LTD indicates how much feedback information helps improve future decisions. It quantifies the \emph{value of information} in the context of dynamic programming and optimal control. 

3) \textbf{CLS}: For each node $i$ belonging to the mitigation campaign, we compute its closeness centrality $cent_i = \frac{1}{\sum_j dis(i,j)}$, where $dis$ is the shortest distance from $i$ to $j$. Then, we assign the budget such that $u_i \propto cent_i$, meaning that budget is distributed to mitigation nodes based on their proximity to nodes according to network structure. Closeness Centrality (CLS) has been widely used in the literature as a baseline for finding influential nodes~\cite{chen2012identifying,de2014role,gao2015identifying}.

4) \textbf{EXP}: The CLS baseline is only structural and does not use the fake news infection data. EXP augments it by computing an Exposure-based Closeness Centrality, $cent^k_i = \sum_j \frac{\sum_{l=1}^L \Fcal^{k-l}_{j}}{dis(i,j)}$, where the numerator is the total number of times node $j$ has been exposed to the fake news campaign in the $L$ intervals before stage $k$. The more times node $j$ has been exposed to the fake news, the more important it is for the mitigation campaign to reach it. EXP assigns the budget according to $u^k_i \propto cent^{k}_i$.

5) \textbf{RND}: This policy assigns a random solution in the convex space of feasible interventions. It serves as a baseline and improvement over this random policy makes comparison feasible across different settings.

\subsection{Synthetic Experiments}
\label{sec:synthetic_experiments}
\xhdr{Setup}
For all except the experiment over network size, the networks were generated synthetically with $n=300$ nodes.
Endogenous intensity coefficients were set as $a_{ij} \sim \Ucal[0, 0.5]$. 
To mimic real world networks, sparsity was set to $0.02$, \emph{i.e.}, each edge was kept with probability 0.02.
The influence matrix was scaled appropriately such that the spectral radius is a random number smaller than one to ensure the stability of the process. 
The Hawkes kernel parameter was set to $\omega = 1$, which means loosing roughly 63 \% of influence after 1 time unit (minutes, hours, etc). 
Both fake news and mitigation processes obey these network settings. Among $n$ nodes, we assume 20 nodes create fake news and another 20 nodes can be incentivized (via the exogenous intensity) to spread true news.
Each stage has length of $\Delta_T = 1$.
The discount factor was set to $\gamma=0.7$.
For determining features, we set $L=2$ and we choose $\Delta_f = \Delta_T$ for simplicity.
The upper bound for the intervention intensity was chosen by $\alpha_i \sim \Ucal[0,0.5]$. 
The price of each person was $c^k_i =1$, and the total budget at stage $k$ was randomly generated as $C_k \sim (n \times \Ucal[0, 0.5])$. 
1000 randomly sampled states were used for the LSTD algorithm.
To evaluate a policy (learned by an algorithm) we simulated the network under that policy 50 times and took the discounted total reward averaged over these 50 runs as an empirical valuation of the policy. 
Furthermore, each single run was simulated for 10 consecutive stages; from the eleventh stage onward, the objectives contribute 0.02 of the total reward and can be discarded. 
For all experiments, the above settings are assumed unless it is explicitly mentioned otherwise.

\begin{figure*}[t]
\centering
  \begin{tabular}{cccc}
  \includegraphics[width=0.24\textwidth]{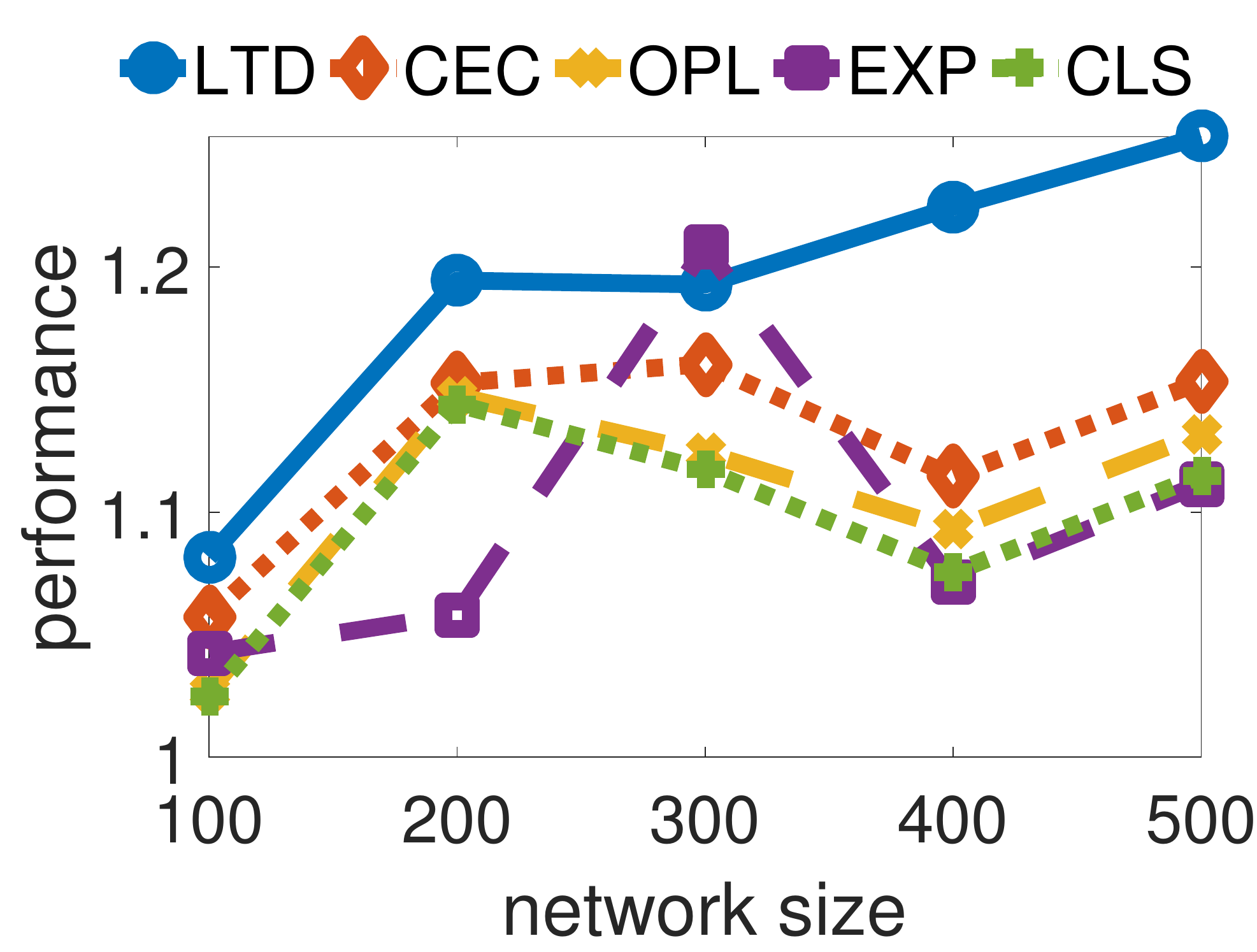}  \hspace{-3mm} &
  \includegraphics[width=0.24\textwidth]{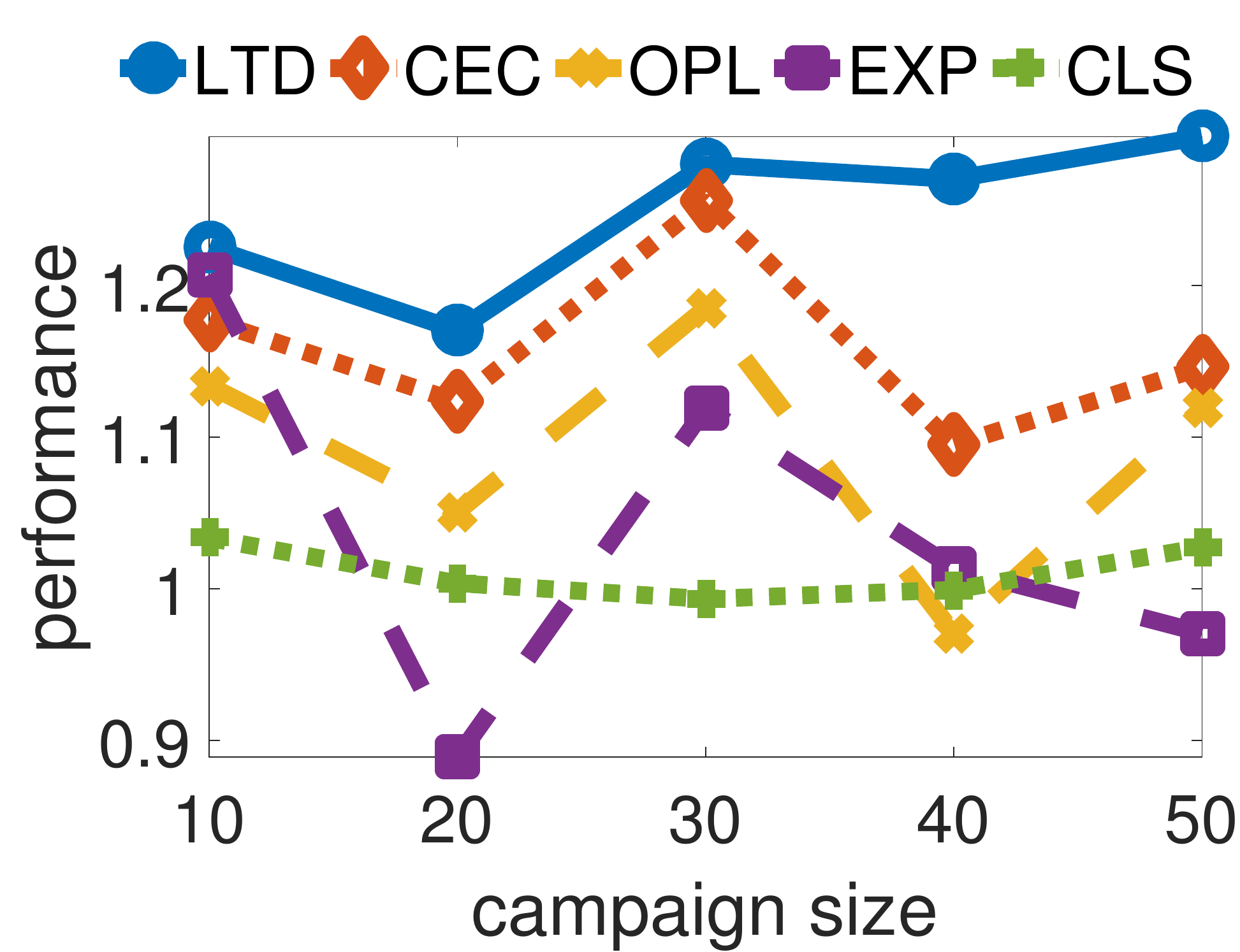} \hspace{-3mm} &
  \includegraphics[width=0.24\textwidth]{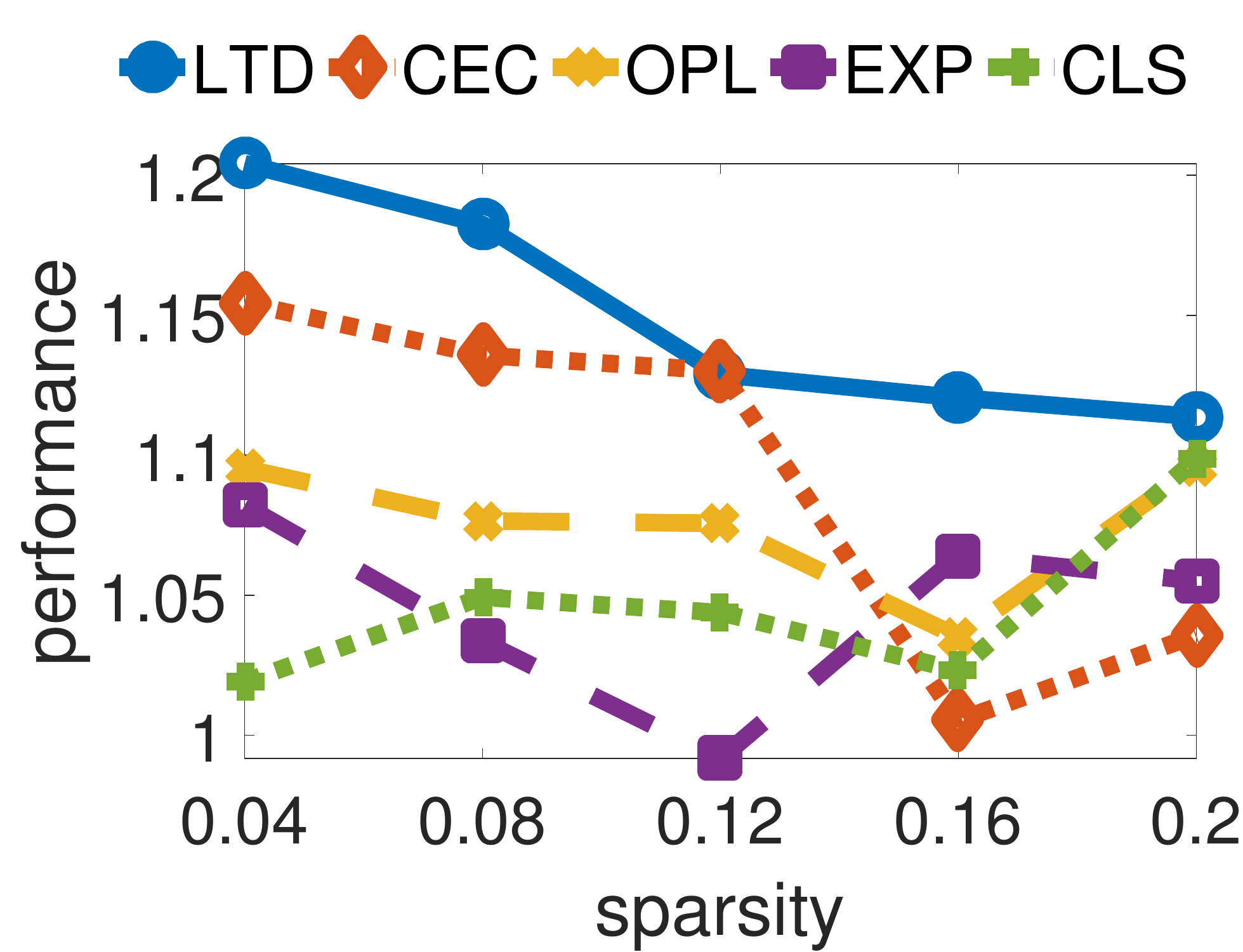} \hspace{-3mm} &
  \includegraphics[width=0.24\textwidth]{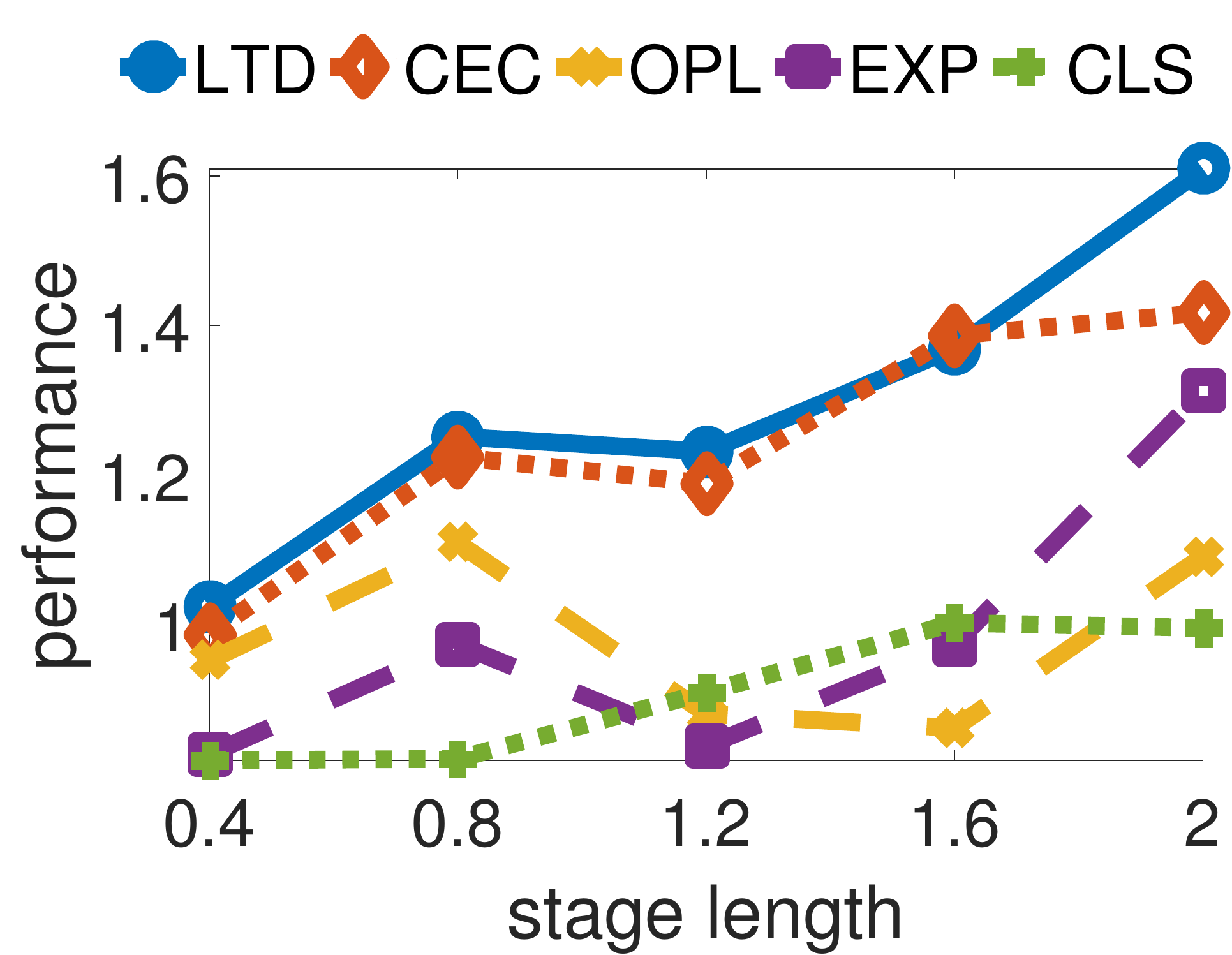}  \hspace{-3mm} \\
  (a) &
  (b) &
  (c) &
  (d) 
  \end{tabular}
  \caption{Performance improvement of different methods over the random policy on synthetic networks for correlation maximization}
  ~\label{fig:cor-synth-results}
\end{figure*}

\begin{figure*}[t]
\centering
  \begin{tabular}{cccc}
   \includegraphics[width=0.24\textwidth]{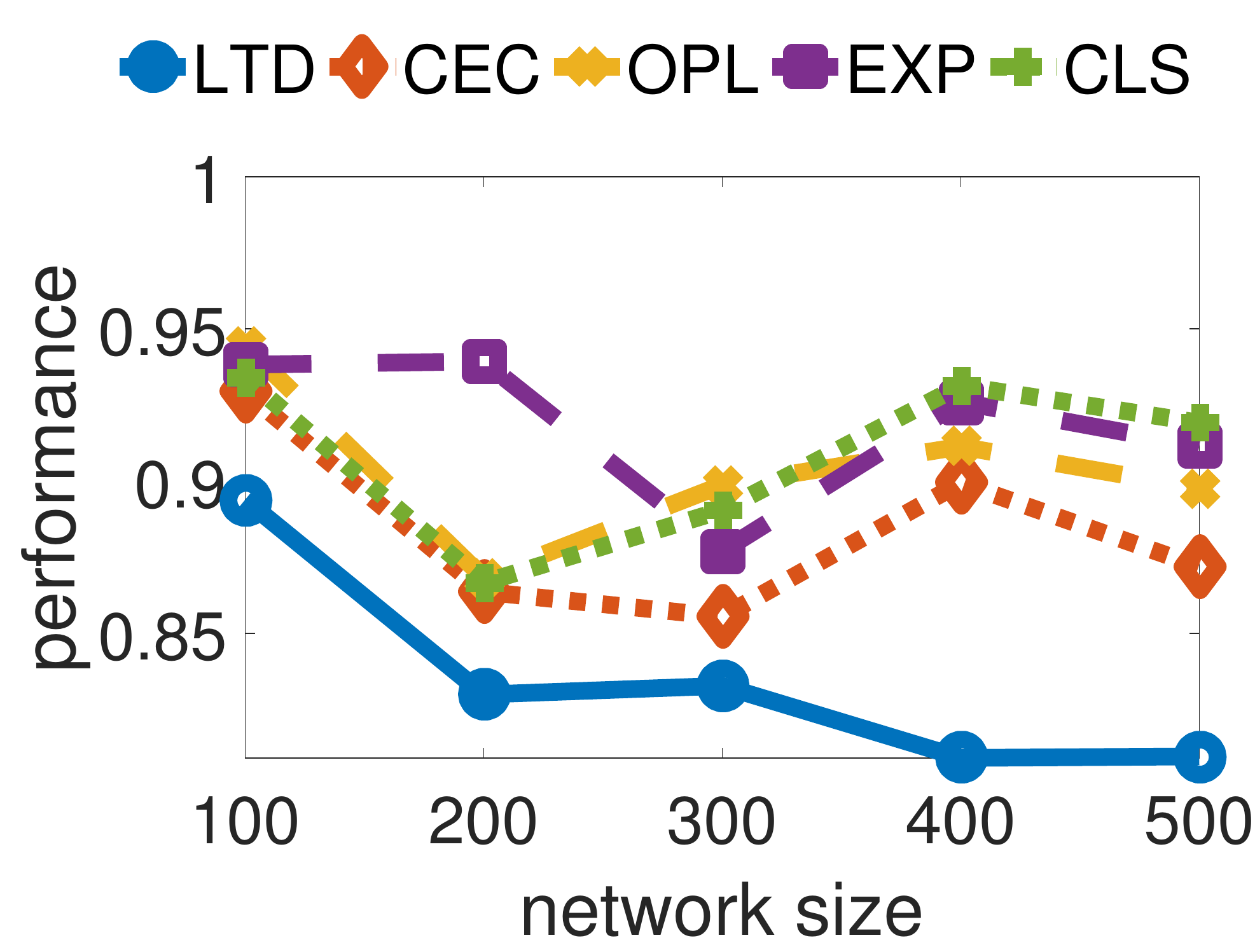}  \hspace{-3mm} &
  \includegraphics[width=0.24\textwidth]{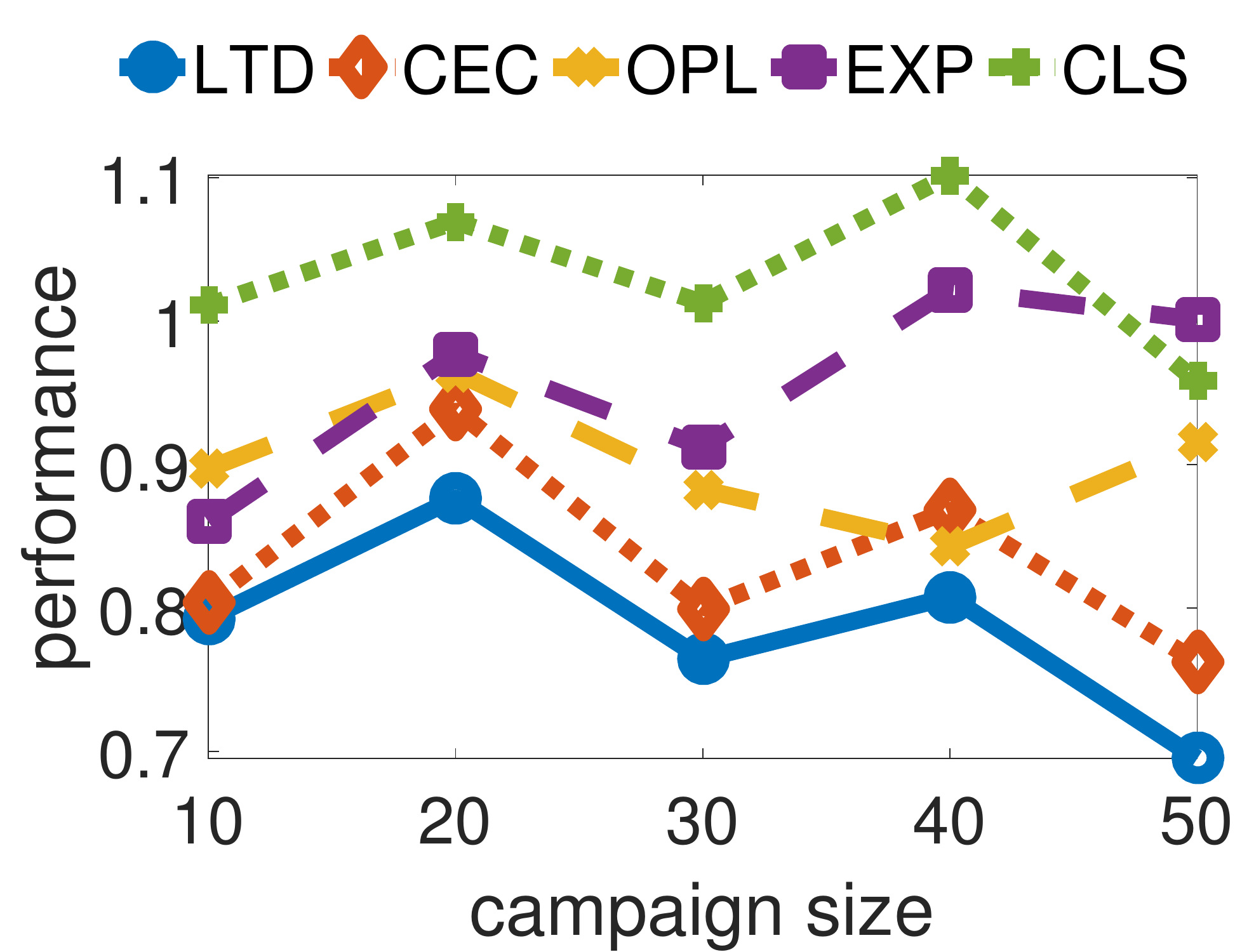} \hspace{-3mm} &
  \includegraphics[width=0.24\textwidth]{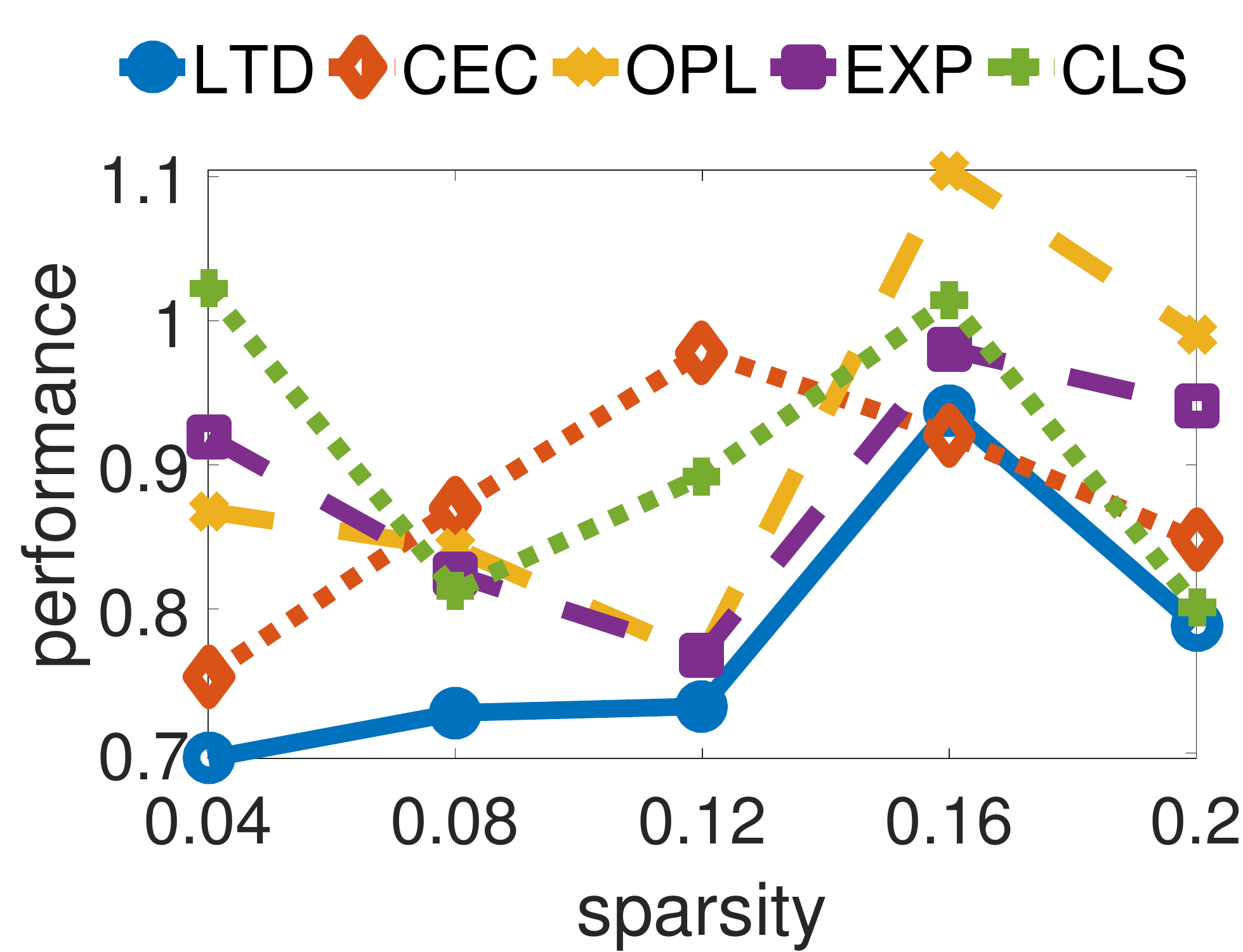} \hspace{-3mm} &
  \includegraphics[width=0.24\textwidth]{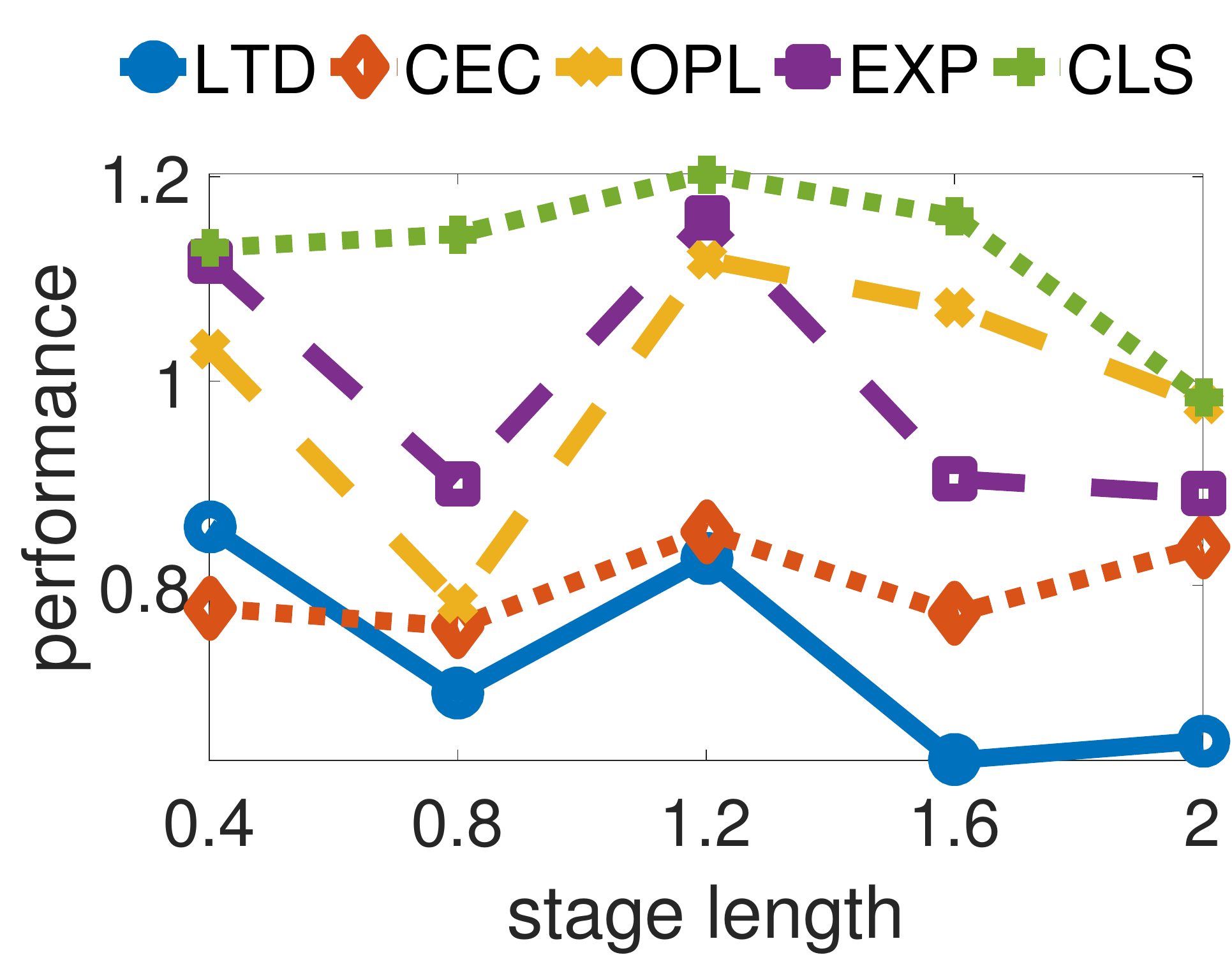}  \hspace{-3mm} \\
  (a) &
  (b) &
  (c) &
  (d) 
  \end{tabular}
  \caption{Performance improvement of different methods over the random policy on synthetic networks for distance minimization}
  ~\label{fig:dis-synth-results}
\end{figure*}

\xhdr{Intervention results}
Fig.~\ref{fig:cor-synth-results} demonstrates the performance of different methods. Performance of a policy is quantified as the ratio of the total reward achieved by running the policy, over the total reward achieved by the random policy (RND). This allows us to compare the effectiveness of the algorithms over a variety of settings. All the results reported are averages over 10 runs with random networks generated according to the above setup. 
Overall, it is clear that LTD is almost consistently the best. It improves over the random policy by roughly 20 percent. 
CEC is the second best and shows the effectiveness of multi-stage and closed loop intervention. 
This validates our intuition that although CEC computes the reward from both fake news and mitigation processes, the lack of explicit features corresponding to previous events in its value function prevents it from learning the reason for the reward.
Roughly, OPL is the third best algorithm, due to its negligence of the state and the actual events that occurred.
Next, comes the EXP algorithm followed by the CLS. The poor performance of these (compared to others) shows that structural properties are not sufficient to tackle the fake news mitigation problem. EXP is roughly better than CEC because it heuristically takes into account the fake news exposure.    

Fig.~\ref{fig:cor-synth-results}-a shows the performance with respect to increasing network size. 
The difference between alternative methods and the gap between LTD and others increase with the network size. Furthermore, the performance of all methods show an increase over random policy when the problem size gets larger. This illustrates 
the fact that efficient distribution of budget matters more when confronted with problems of increasing complexity and size.

Fig.~\ref{fig:cor-synth-results}-b shows the performance with respect to increasing the mitigation campaign size. Larger campaigns imply greater flexibility of intervention, which can be exploited by clever algorithms to achieve higher performance.

Fig.~\ref{fig:cor-synth-results}-c shows the performance with respect to increasing sparsity of the network. Interestingly, the performance of all the algorithms move towards to the random policy as the network becomes denser. This can be understood by considering a complete graph, so that no  matter how and to whom we distribute the mitigation budget, all the nodes are exposed to the mitigation campaign almost equally. However, since real social networks are usually sparse, the effectiveness of the proposed method stands out.

Finally, Fig.~\ref{fig:cor-synth-results}-d shows the performance with respect to the length of an stage. Longer stage lengths increase the potential for a good policy to attain higher reward than a random policy, and this is reflected by the sharp increase and larger performance gap between LTD and others for longer lengths.
We observe the same patterns for the distance minimization in Fig.~\ref{fig:dis-synth-results} problem and avoid repeating them.

\begin{figure}[t]
\centering
  \begin{tabular}{cc}
   \hspace{-4mm}
  \includegraphics[width=0.24\textwidth]{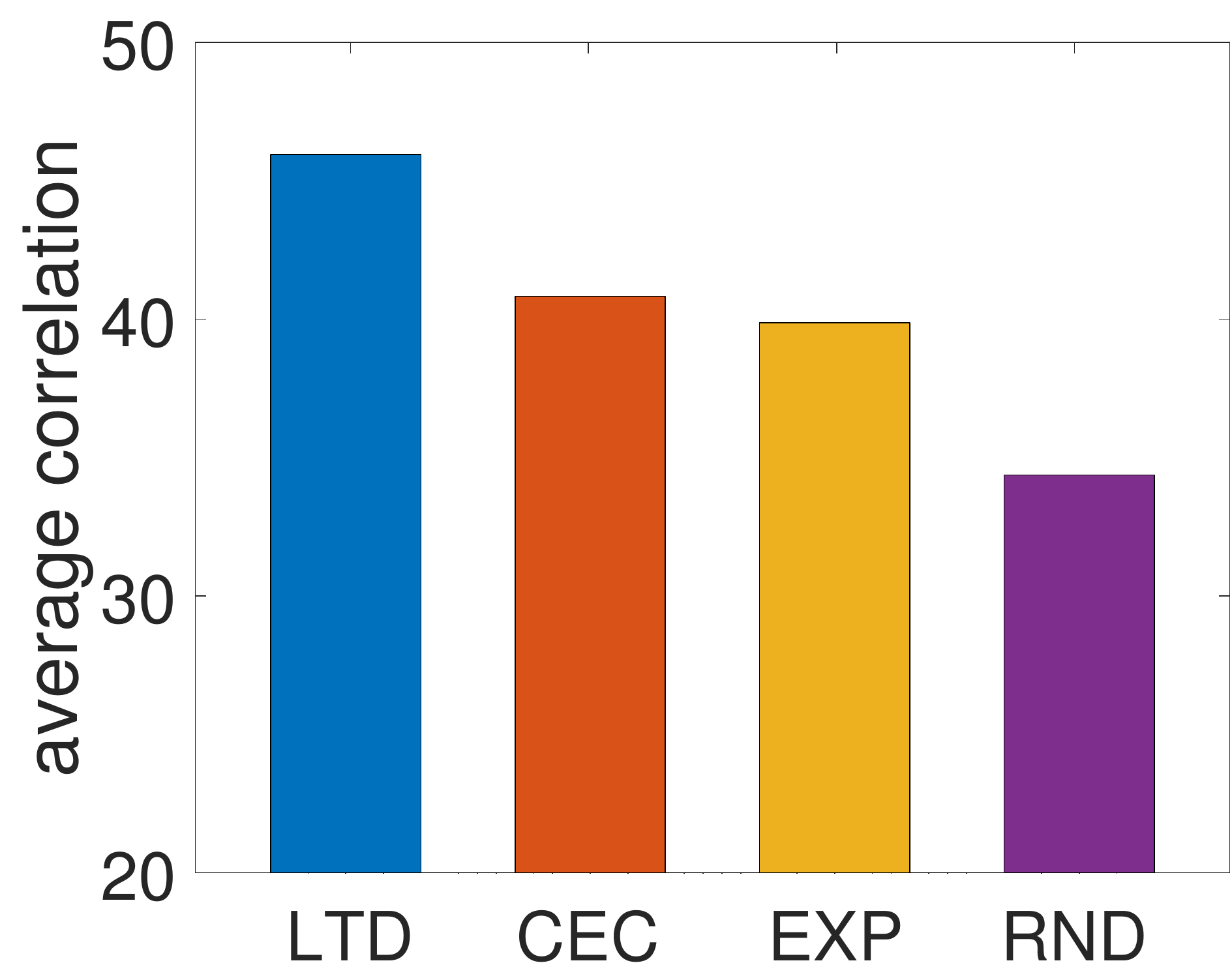} \hspace{-4mm} &
  \includegraphics[width=0.24\textwidth]{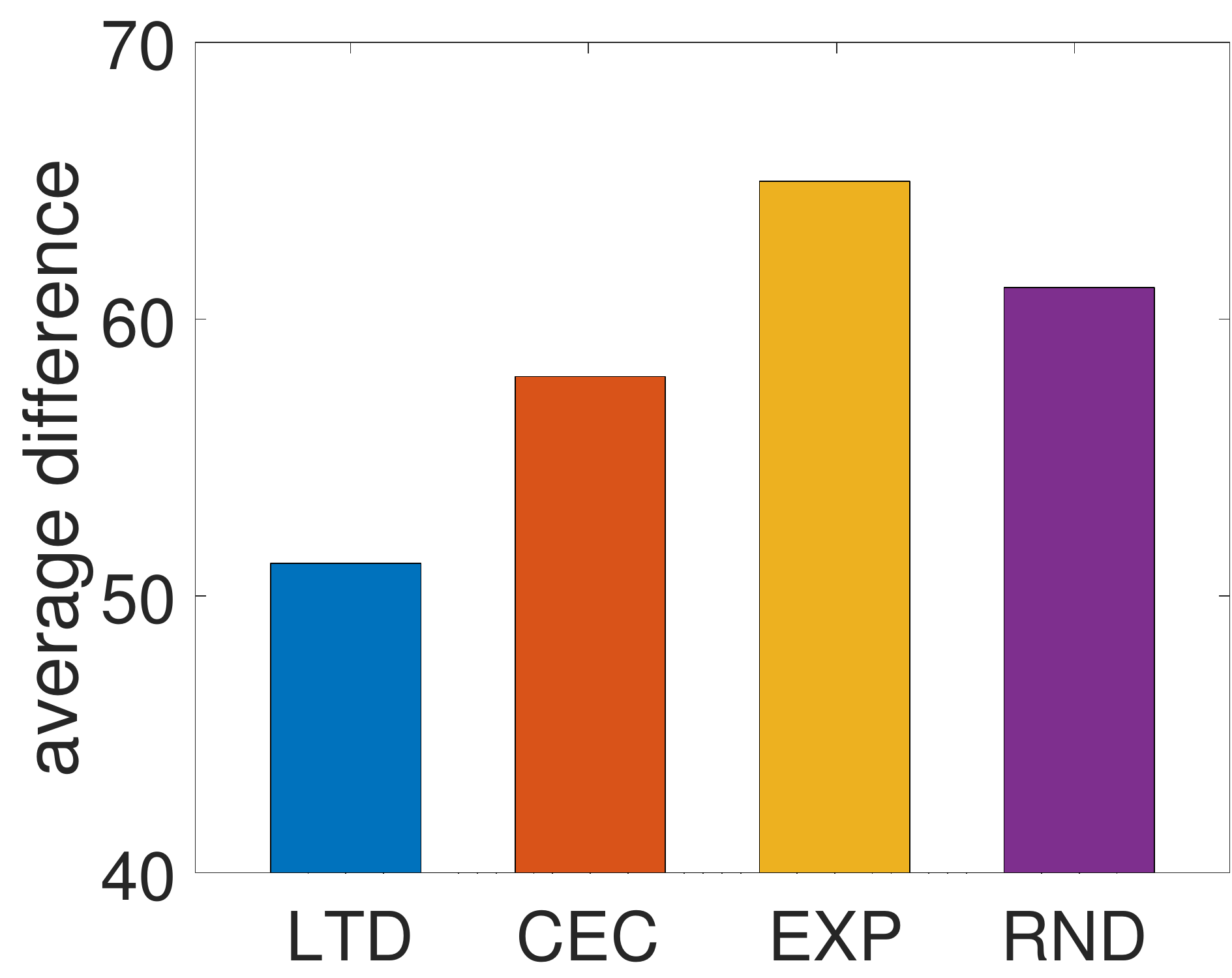}  \\
  (a) Correlation &
  (b) Difference
  \end{tabular}
  \caption{Results of fake news mitigation on Twitter network}
  ~\label{fig:real}
\end{figure}

\subsection{Real experiments}
\label{sec:real_experiments}
In this section we explain our real-time intervention results. To the best of our knowledge, we are the first to employ a real-time experiment to evaluate a point process based social network intervention strategy.

\xhdr{Setup}
Using five Twitter accounts, each of which made five posts on machine learning topics at random times per day for a span of two months (Nov.-Dec. 2016), we accumulated a network of 1894 real users with 23407 directed edges in total. 
We used this historical data to learn the network parameters $\lbrace \alpha_{ij}, \mu^i \rbrace$ using maximum likelihood (similar to related work~\cite{zhou2013learning,Farajtabar:2014a}) with one hour as the time resolution and the kernel decay parameter $\omega$ set to 0.1.
As illustrated in Fig.\ref{fig:blockdiagram} the optimal policy was learned using LSTD and policy improvement. 
Then the real-time experiment starts:
Two of the accounts, interpreted as the source of fake news, continued to behave using the same randomized policy as they did in the data collection stage, while the posting times of the other three accounts were generated from $(u_1, u_2, u_3)^T$, produced by our LTD strategy or a competitor strategy.
Each policy was run for 10 stages of length 12 hours. Therefore, $\- \Delta_T = \Delta_f=12$. Since both fake news and mitigation accounts were tweeting random posts on machine learning, we assume negligible bias in the content that can confound the performance. 
At the end of each stage, all retweets--by users within the network--of the posts made during the two most recent stages were used to construct the feature vector and compute the value function, which was used to find the optimal intervention for the next stage. The methods CEC and  OPL belong to the same category, and it has been shown that CEC outperforms OPL in ~\cite{Farajtabar:2016b}. Furthermore,  EXP and CLS also belong to similar families and our synthetic experiments confirm the superiority of the former. So, to save time in real interventions, we only test CEC from the first and EXP from the second pair, and compare them with the random policy (RND) and with our algorithm (LTD).

\xhdr{Real-time intervention results}
Fig.~\ref{fig:real} shows the performance of our results compared to competitors. The results show that our approach outperforms the other three baselines by a reasonable margin. As expected CEC is the second best algorithm with a margin of 5 for the correlation maximization objective. It translates to increase in amount of correlation equal to 5, which is a noticeable amount. Furthermore, in the difference minimization task, our approach reached around 7 in difference. This means that we decreased the difference in exposure to the two processes to less 2.6 per user, which is considerable improvement. 
For both tasks, LTD made more mitigation posts over all daytime phases than it did over all nighttime phases, whereas the competitor strategies did the opposite. This could be a reason for its better performance.
One surprising fact is that the number of retweets by users outside the network, which was not used for our features, can exceed the number of retweets by users within the network. This is because the ``hashtag" feature on Twitter allows posts to be seen by a much larger set of users, who do not necessarily follow the source accounts. In addition to retweets, users can also ``like" a post, indicating that they were exposed to fake or real news; while we measured this, we did not include it in the reward. Future experiments can use these two observations to widen the experimental scope and more accurately measure the effectiveness of a mitigation strategy. 
Despite having these limitations, our experiment 
serves as a proof-of-concept for the applicability of point process based intervention in networks, and--to the best of our knowledge--is the first to verify the superiority of a method in a real-time, real-world intervention setting.

\begin{figure}[t]
\centering
  \begin{tabular}{cc}
   \hspace{-4mm}
  \includegraphics[width=0.24\textwidth]{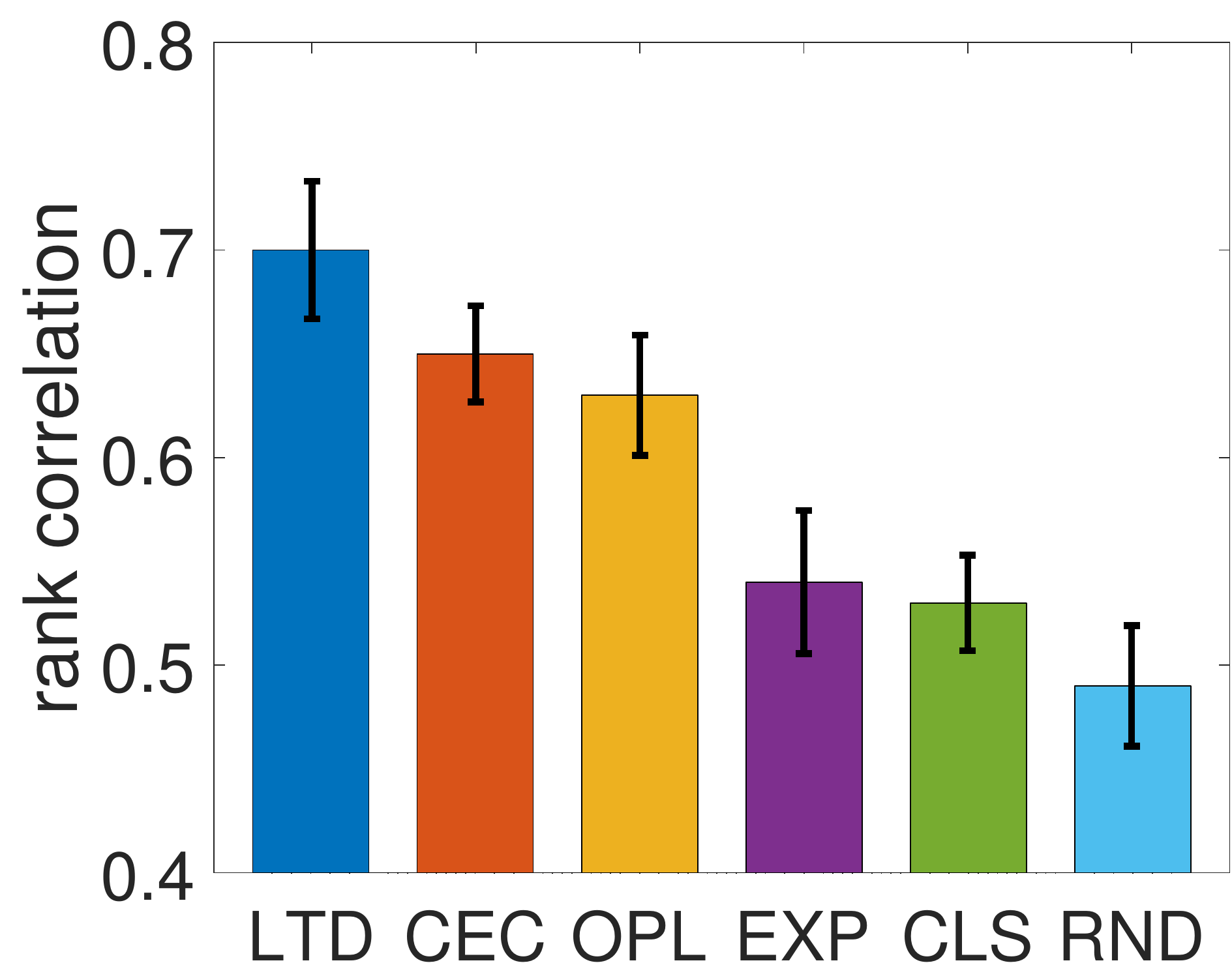} \hspace{-4mm} &
  \includegraphics[width=0.24\textwidth]{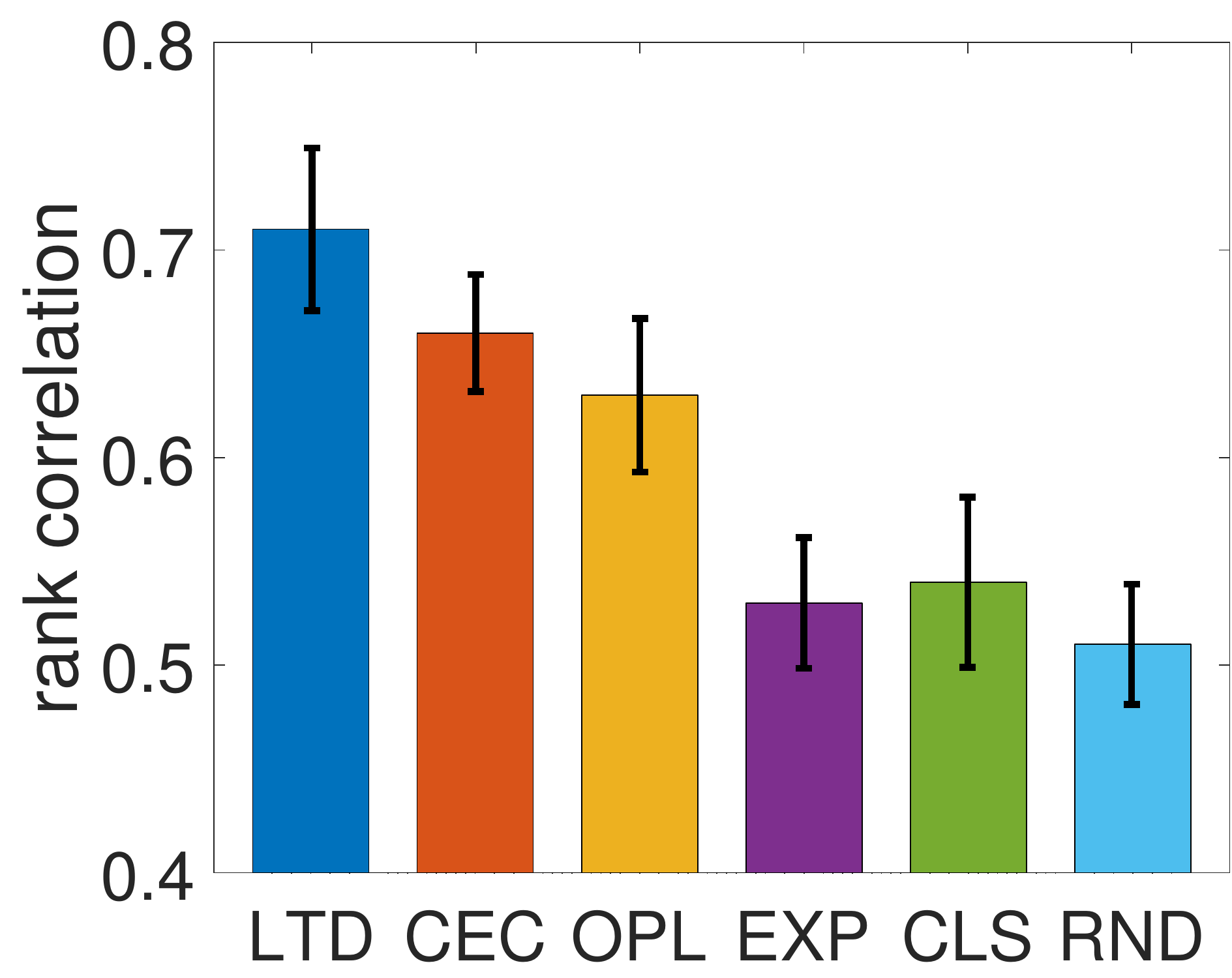}  \\
  (a) Correlation &
  (b) Difference
  \end{tabular}
  \caption{Rank correlation for prediction}
  ~\label{fig:ranks}
\end{figure}

\xhdr{Prediction evaluation results}
The previous part described the evaluation scheme of real-time intervention in a social media platform. 
In this part, we used historical real data to mimic this procedure.
We extracted 12 full 10-stage trajectories of events from the 2-month historical data under the random policy. %
For any of these 10 pairs, the methods were evaluated according to how well they predict the relative ordering among these 12 trajectories (with respect to the objective function). 
To evaluate each method, we created a sorted list of these 12 trajectories according to increasing objective, and created a second list sorted by increasing closeness to the intervention method.
This closeness is the mean squared error between the prescribed intervention and actual intensity, which we inferred using maximum likelihood.
Then, by computing the rank correlation of the two sorted lists, and repeating for each of the five methods, we can find out how well they perform on the prediction task. 
A better predictor is expected to be a better mitigation strategy.
Fig.~\ref{fig:ranks} shows the performance.  
%
%

%
%
%

%
%

\begin{figure}[t]
\centering
  \begin{tabular}{cc}
   \hspace{-4mm}
  \includegraphics[width=0.24\textwidth]{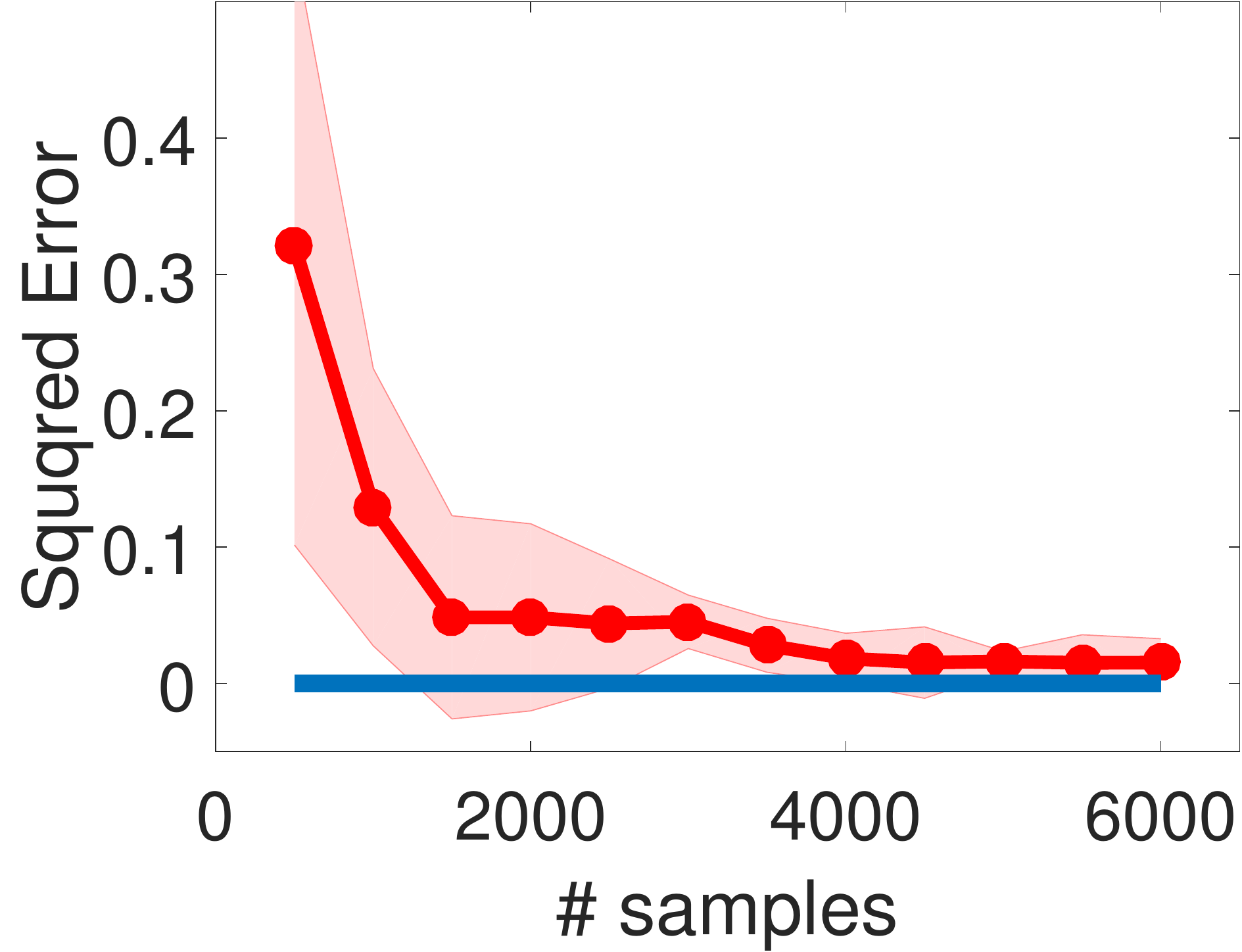} \hspace{-4mm} &
  \includegraphics[width=0.24\textwidth]{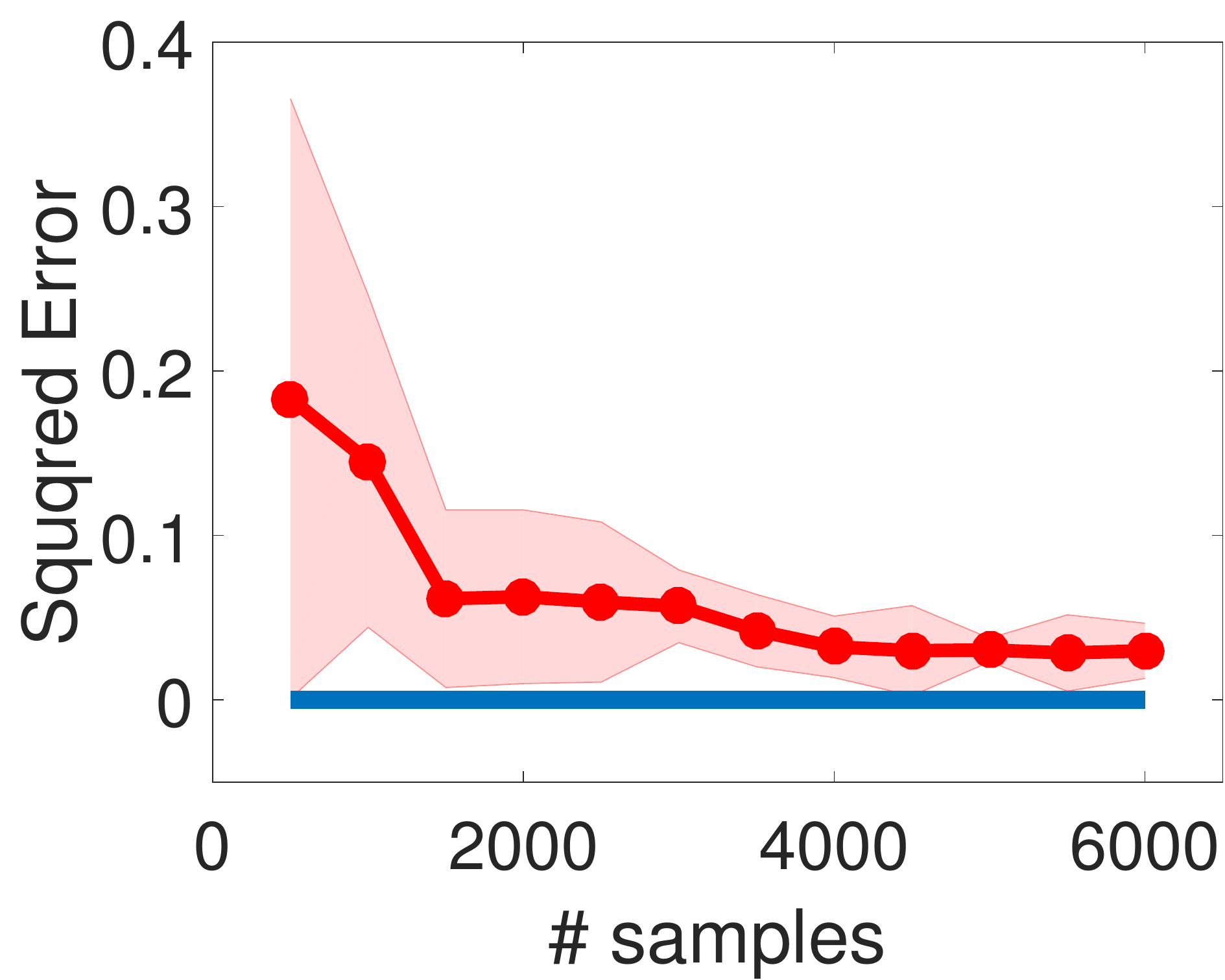}  \\
  (a) Correlation &
  (b) Difference
  \end{tabular}
  \caption{Convergence of linear approximated value function}
  ~\label{fig:convergence}
\end{figure}

\subsection{Linear approximation accuracy}
\label{sec:linear-conv}
In our LSTD algorithm we used a linear approximation for the value-function. One might ask how accurately can linear features approximate the value-function. To this end, we take a sample state $x$ as a state with no prior activity and intensity $\psi(x) = (0, \dots, 0, 1)$.
This is the the initial state assumed in our experiment runs. First we empirically found $V^{\pi}(x)$ under the learned policy by simulating the process 100 times each with 10 stages. We compare the empirical average and the standard deviation of the total reward with the one estimated by the linear approximation $\psi^{\top} w^{\pi}$. Fig.~\ref{fig:convergence} shows the results for correlation maximization and difference minimization. In both cases, by increasing the number of samples (used in LSTD), the estimated $w$ leads to better estimation of $V^{\pi}(x)$. 
First, the figure shows that we can achieve a reasonable accuracy with a fair amount of samples. 
Secondly, although it appears that the approximation is converging to the empirical value, we notice that increasing the number of samples beyond 4000 does not improve the error, which maintains a constant distance from 0. We believe this is because the optimal value function does not lie in the linear span of the feature space. Employing more complex features, such as polynomial features and deep neural network based representations, remain as interesting avenues for future work.

\section{Discussion}
\label{sec:discus}
The use of point process based intervention comes with a tradeoff: 
on one hand, the stochastic nature of multivariate point processes allows it to model the uncertainty of event occurrences in real-world networks;
however, by adopting this stochastic model, the intervention policy can only set the optimal conditional intensity, rather than the precise best times, of fake news mitigation events.
This can be improved in future experiments by choosing shorter time intervals for stages in the mitigation campaign.
An assumption made in the real-world experiment is that all events by user $u$ are seen by user $v$ if $v$ follows $u$, meaning that fake or real news events at $u$ are seen by $v$. 
While it is true for Twitter that all tweets by $u$ will appear on the home timeline of $v$ who follows $u$ \cite{Twitter:2016a}, it is not necessarily true that a follower $v$ will see these tweets (suppose they did not access Twitter that day). 
Therefore, future experiments can improve the accuracy of reward and performance measurements by estimating the probability of users being online during certain time intervals and seeing tweets from accounts they follow. 

It is important to note that Òmitigating fake newsÓ is a vague and qualitative goal, and it does not necessarily imply a reduction of fake news events. 
Matching the exposure of users to real and fake news is one of many possible objectives for arriving at a precise quantitative realization of Òmitigating fake newsÓ.  
One can define any objective function that accounts for the number of exposures or events from fake and real news cascades. 
Furthermore, the ÒexposureÓ itself can be interpreted in many different ways depending on the application. 
However, whenever there are multiple dependent event sequences and the rate function of one or many can be controlled, the point process framework naturally allows one to define objective functions based on events and to find an optimal policy with respect to the desired goal.

We acknowledge that the experiments conducted in this present work do not directly test for a reduction of fake news events, and that the content-neutral real-time experiment does not perfectly represent fake news processes, as semantic content may also contribute to propagation dynamics. 
From a modeling standpoint, introducing interactions between the fake and real news processes may allow one to design objectives that specifically reward the reduction of fake news events.
Aside from subjecting real users to actual fake news, more complex real-world experiments that are not content-neutral, involving two competing opinions, may allow one to better measure the effectiveness of various methods in reducing the spread of one opinion.
Furthermore, for future work we would like to incorporate more complex features, such as quadratic and nonlinear features. 
Utilizing deep neural nets is also an interesting future work for modeling complex feature set.


{\bf Acknowledgement.}
This project is supported in part by NSF IIS-1639792, CNS-1409635, NSF DMS-1620342, NSF IIS-1218749, NIH BIGDATA 1R01GM108341, NSF CAREER IIS-1350983, ONR N00014-15-1-2340, NVIDIA, Intel, and Amazon AWS.


\bibliographystyle{icml2017}
\bibliography{library,library2}

\newpage
\appendix

\section{Proof of Theorem \ref{thm:secondorder}}
\begin{proof}
Fix node index $j$ and $t'\geq0$, define $g_{ji}(t',t)$ for all node $i$ and $t$ such that
\begin{equation}\label{eq:gij}
g_{ji}(t',t)\dif t = \EE \sbr{\dif N_i(t)|\dif N_j(t')=1} - \delta_{ij} \delta(t-t') \dif t- \eta_i(t) \dif t
\end{equation}
Since the conditional intensity of $N_i(t)$ is $\lambda_i(t)$, we have
\begin{align*}
g_{ji}(t',t)\dif t & = \EE \sbr{\dif N_i(t)|\dif N_j(t')=1}-\delta_{ij}\delta(t-t')\dif t-\eta_i(t)\dif t \\
& = \EE\sbr{\lambda_i(t)|\dif N_j(t')=1}\dif t-\delta_{ij}\delta(t-t')\dif t-\eta_i(t)\dif t
\end{align*}
Furthermore, we have $\lambda_i(t)=\mu_i(t)+ \sum_{k=1}^{n}\int_0^t \phi_{ki}(t-s)\dif N_k(s)$ and hence
\begin{align*}
\EE\sbr{\lambda_i(t)|\dif N_j(t')=1}
& = \mu_i(t) + \sum_{k=1}^{n}\int_0^t\phi_{ki}(t-s)\EE[\dif N_k(s)|\dif N_j(t')=1]\\
& = \mu_i(t) + \sum_{k=1}^{n}\int_0^t\phi_{ki}(t-s)\sbr{g_{jk}(t',s)\dif s+\delta_{kj}\delta(s-t')\dif s+\eta_k(s)\dif s}\\
& = \mu_i(t) + \sum_{k=1}^{n}\int_0^t\phi_{ki}(t-s)g_{jk}(t',s)\dif s + \phi_{ji}(t-t')+
\sum_{k=1}^n\int_0^t\phi_{ki}(t-s)\eta_k(s)\dif s
\end{align*}
where we applied the definition of $g_{jk}$ in \eqref{eq:gij} to obtain the second equality.
Combining the two equations above and using the fact that 
$\eta_i(t)=\mu_i(t)+\sum_{k=1}^n\int_0^t\phi_{ki}(t-s)\eta_k(s)\dif s$,
we obtain that
\begin{align*}
g_{ji}(t',t)
& = \sum_{k=1}^{n}\int_0^t\phi_{ki}(t-s)g_{jk}(t',s)\dif s+\phi_{ji}(t-t')-\delta_{ij}\delta(t-t')
\end{align*}
Since $j$ and $t'$ are arbitrary, 
we let $G(t',t)$ be the matrix such that the $(j,i)$-th entry of $G(t',t)$ is $g_{ji}(t',t)$, then we have
\begin{equation}\label{eq:WH_G}
G(t',t)=G(t',t) \ast \Phi(t) + \Phi(t-t') - \delta(t-t')I
\end{equation}
Note that the Wiener-Hopf equation \eqref{eq:WH_G} determines the unique solution $G(t',t)$ for 
all $t\geq t'$.
Moreover, since MHP is simple and that $\dif N_i(t)=0$ or $1$ a.s. for all $i$, we have
\begin{align}
\EE[\dif N_i(t)\dif N_j(t')] 
&= \pr \del{\dif N_i(t)=1,\dif N_j(t')=1} \nonumber\\
&= \pr \del{\dif N_i(t)|\dif N_j(t')=1}\pr\del{\dif N_j(t')=1} \nonumber\\
&= \EE[\dif N_i(t)|\dif N_j(t)=1] \EE[\dif N_j(t')] \label{eq:dNidNj}\\
&= \EE[\dif N_i(t)|\dif N_j(t)=1] \EE[\lambda_j(t')]\dif t' \nonumber \\
&= \EE[\dif N_i(t)|\dif N_j(t)=1]\eta_j(t')\dif t' \nonumber \\
&=g_{ji}(t',t)\eta_j(t')\dif t\dif t' + \delta_{ij}\delta(t-t')\eta_j(t')\dif t\dif t' + \eta_i(t)\eta_j(t')\dif t\dif t' \nonumber
\end{align}
Similarly, we can switch $i$ and $j$, and $t$ and $t'$ to obtain
\begin{equation}\label{eq:dNjdNi}
\EE[\dif N_i(t)\dif N_j(t')] 
=g_{ij}(t,t')\eta_i(t)\dif t\dif t' + \delta_{ij}\delta(t-t')\eta_i(t)\dif t\dif t' + \eta_i(t)\eta_j(t')\dif t\dif t'
\end{equation}
Combining  \eqref{eq:dNidNj} and \eqref{eq:dNjdNi} we have that
$$ g_{ji}(t',t)\eta_j(t')=g_{ij}(t,t')\eta_i(t)$$
i.e., $G(t',t)\trans \Sigma(t')=\Sigma(t)G(t,t')$,
from which $G(t',t)$ for $t<t'$ is also uniquely determined. We therefore have
\begin{equation}
\EE \sbr{\dif N(t)\dif N(t')\trans  } = G(t',t)\trans \Sigma(t')\dif t\dif t' + \delta(t-t')\Sigma(t')\dif t\dif t'+\eta(t)\eta(t')\trans \dif t\dif t'
\end{equation}
This completes the proof.
\end{proof}

\section{Details of Policy Evaluation}
\label{sec:policy-eval}
We seek an approximate value function $\hat{v}^{\pi}$ that is invariant under one application of the Bellman operator $T^{\pi} $ followed by orthogonal projection:
\begin{align}
\hat{v}^{\pi}  = \Psi(\Psi^{\top}\Psi)^{-1}\Psi^{\top} (T^{\pi}   \hat{v}^{\pi} )  
\end{align}
By replacing the linear approximation, $\Psi  w^{\pi} = v^{\pi}$ ,  and  some manipulations we get:
\begin{align*}
 \Psi(\Psi^{\top}\Psi)^{-1}\Psi^{\top}  (r^{\pi}   + \gamma \Psi' w^{\pi} ) &=  \Psi  w^{\pi} \\
 \Psi \left( (\Psi^{\top}\Psi)^{-1}\Psi^{\top}  ( r^{\pi}   + \gamma \Psi' w^{\pi} ) - w^{\pi}  \right)   & = 0 \\
 (\Psi^{\top}\Psi)^{-1}\Psi^{\top}  ( r^{\pi}   + \gamma \Psi' w^{\pi} ) - w^{\pi}   & = 0 \\
  (\Psi^{\top}\Psi)^{-1}\Psi^{\top}  ( r^{\pi}   + \gamma \Psi' w^{\pi} ) & = w^{\pi}  \\
 \Psi^{\top}  ( r^{\pi}   + \gamma \Psi' w^{\pi} ) & =  (\Psi^{\top}\Psi) w^{\pi}  \\ 
 \underbrace{ \Psi^{\top}  (  \Psi   - \gamma \Psi')}_{D\times D}  w^{\pi}  & =  \underbrace{\Psi^{\top} r^{\pi}}_{D \times 1 }    
\end{align*}
Defining $A^{\pi} = \Psi^{\top}  (  \Psi   - \gamma \Psi')$ and $b^{\pi} =  \Psi^{\top} r^{\pi}$ the estimated coefficients are the solution of a $D \times D$ linear systems of equation: $A^{\pi} \omega^{\pi} = b^{\pi}$.


\section{Details of Policy Improvement}
\label{sec:app:policy-improve}
Assume we are at the beginning of stage $k$.The expected feature vector for the next state $x'$ is comprised of $L$ intervals per process, out of which $L-1$ are observed. Only the most recent interval is not observed and needs to be re-evaluated in expectation sense. 
To compute $\EE[V^{\pi}(x')]$ have:
\begin{align}
\EE[V^{\pi}(x')]  = & \EE[\sum_{d=1}^D w^{\pi}_d \psi_d(x')]  \\
 = & \EE[
\sum_{i=1\ldots n,l=1\ldots L} w^{\pi}_{(l-1)n+i} z^{k}_{M,(l-1)n+i} 
+ w^{\pi}_{nL+(l-1)n+i} z^k_{F,(l-1)n+i} 
+ w^{\pi}_{2nL+1} ]
\\ 
 = &
\sum_{i=1\ldots n,l=1\ldots L-1} w^{\pi}_{ln+i} z^{k-1}_{M,(l-1)n+i} 
+ w^{\pi}_{nL+ln+i} z^{k-1}_{F,(l-1)n+i}
+ \sum_{i=1\ldots n} w_{i} \EE[z^k_{M,i}]+ w_{nL+i} \EE[z^k_{F,i}] \\
& + w^{\pi}_{2nL+1} 
\end{align}

Then, following~\cite{Farajtabar:2016b}, we obtain
\begin{align}
\EE[z^{k}_{M}] = \Gamma \, (\mu^M + u^k) + \Upsilon \, y^k_M \\
\EE[z^{k}_{F}] = \Gamma \, \mu^F + \Upsilon \, y^k_M 
\end{align}
where 
\begin{align}
& \Upsilon =  (A-\omega I)^{-1} (e^{ (A-\omega I) \Delta}- I) \\
& \Gamma =   \Upsilon + (A-\omega I)^{-1} (\Upsilon- I\Delta)/\omega;
\end{align}

To find $\EE[R(x, u)]$ for the two different reward functions we have defined

\begin{itemize}
\item Correlation Maximization
\begin{align}
\EE[R(x^k, u^k)] = &
\frac{1}{n} \EE[ \Mcal^k(\tau_{k+1};x^k,u^k)^{\top} \Fcal^k(\tau_{k+1};x^k,u^k) ]
 = 
\frac{1}{n}\EE[ {z^k_{M}}^{\top} B^{\top} B \, z^k_{F} ]  \\
= & \frac{1}{n} \EE[ {z^k_{M}}]^{\top} \, B^{\top}  B  \, \EE[z^k_{F} ] 
= \frac{1}{n} \left( \Gamma (\mu^M + u^k ) + \Upsilon y^k_M \right)^{\top} B^{\top} B 
\left( \Gamma \mu^F  + \Upsilon y^k_F \right)
\end{align}
Here the second line is due to the fact that mitigation campaign and fake news process are independent of each other (given the network model).
Note the linear dependence of the the objective on our intervention $u^k$ which combined with linear constraints result in a convex optimization problem.
\item Difference Minimization
\begin{align}
\EE [ R(x^k,u^k)] =  & -\frac{1}{n} \EE[ \, \enVert{\Mcal^k(\tau_{k+1};x^k,u^k)-\Fcal^k(\tau_{k+1};x^k,u^k)}^2 \, ]
\\
= &  
-\frac{1}{n}\EE[ (B z^k_{M} - B \, z^k_{F})^{\top}  (B z^k_{M} - B \, z^k_{F}) ] \\
= &  
-\frac{1}{n} \underbrace{\EE[ {z^k_{M}}^{\top} \, B^{\top}  B  \, z^k_{M} ]}_{\text{Second order moments}} 
+\frac{2}{n} \underbrace{\EE[ {z^k_{M}}]^{\top} \, B^{\top}  B  \, \EE[z^k_{F} ]}_{\text{First order moments}}  
- \frac{1}{n} \underbrace{\EE[ {z^k_{F}}^{\top} \, B^{\top}  B  \, z^k_{F} ]}_{\text{Second order moments}}  
\end{align}
The first and second order moments are computed by \citep{Farajtabar:2014a} and Theorem \ref{thm:secondorder}, respectively. 
\end{itemize}

\end{document}